\documentclass{ieeeaccess}
\usepackage{amsmath,amssymb,amsfonts,bm}
\usepackage{array}
\usepackage{algorithm}
\usepackage{algorithmicx}
\usepackage{algpseudocode}
\usepackage{bm}
\usepackage{color}
\usepackage{cite}
\usepackage{epstopdf}
\usepackage{float}
\usepackage{graphicx}
\usepackage{hyperref}
\usepackage{stmaryrd}
\usepackage{stfloats}
\usepackage{subfigure}
\usepackage{mathtools}
\usepackage{multirow}
\usepackage{multicol}
\usepackage{textcomp}
\usepackage[dvipsnames]{xcolor}
\def\BibTeX{\mathrm{B\kern-.05em{\sc i\kern-.025em b}\kern-.08em
    T\kern-.1667em\lower.7ex\hbox{E}\kern-.125emX}}

\begin{document}

\history{Date of publication xxxx 00, 0000, date of current version xxxx 00, 0000.}
\doi{10.1109/ACCESS.2019.DOI}

\title{Dual Weighted $\ell_p$-norm Minimization with Application in Image Denoising}
\author{\uppercase{Huiwen Dong}\authorrefmark{1},
        \uppercase{Jing Yu}\authorrefmark{1},
        \uppercase{and Chuangbai Xiao}\authorrefmark{1}
      }
\address[1]{Faculty of Information Technology, Beijing University of Technology, Beijing 100124, China.}
\tfootnote{This work was supported in part by the National Natural Science Foundation of China under Grant 61501008
           and Beijing Municipal Natural Science Foundation under Grant 4172002.}

\markboth
{Huiwen Dong \headeretal: Preparation of Papers for IEEE TRANSACTIONS and JOURNALS}
{Huiwen Dong \headeretal: Preparation of Papers for IEEE TRANSACTIONS and JOURNALS}

\corresp{Corresponding author: Chuangbai Xiao (e-mail: cbxiao@bjut.edu.cn).}

\begin{abstract}
The robust principal component analysis (RPCA), which aims to estimate underlying low-rank and sparse structures from the degraded observation data, has found wide applications in computer vision. It is usually replaced by the principal component pursuit (PCP) model in order to pursue the convex property, leading to the undesirable overshrink problem. In this paper, we propose a dual weighted $\ell_p$-norm (DWLP) model with a more reasonable weighting rule and weaker powers, which greatly generalizes the previous work and provides a better approximation to the rank minimization problem for original matrix as well as the $\ell_0$-norm minimization problem for sparse data. Moreover, an approximate closed-form solution is introduced to solve the $\ell_p$-norm minimization, which has more stability in the nonconvex optimization and provides a more accurate estimation for the low-rank and sparse matrix recovery problem. We then apply the DWLP model to remove salt-and-pepper noise by exploiting the image nonlocal self-similarity. Both qualitative and quantitative experiments demonstrate that the proposed method outperforms other state-of-the-art methods. In terms of PSNR evaluation, our DWLP achieves about 7.188dB, 5.078dB, 3.854dB, 2.536dB and 0.158dB improvements over the current WSNM-RPCA under 10\% to 50\% salt-and-pepper noise with an interval 10\% respectively.
\end{abstract}

\begin{keywords}
Image denoising, nonlocal self-similarity, robust principal component analysis (RPCA), low-rank, sparsity.
\end{keywords}

\titlepgskip=-15pt

\maketitle

\section{Introduction}
The principal component analysis (PCA) is widely used in data analysis and dimensionality reduction, which remains optimal in the
low-rank matrix recovery problem when the noise is subject to Gaussian distribution, but non-Gaussian noise, even a single gross error, could deviate the estimation far from its ground truth. In 2009, John and Cand\'{e}s \textit{et al.} \cite{rpcaJohn, rpcaCandes} propose a robust principal component analysis (RPCA) model that solves the data sensitivity problem of PCA by decomposing the observation data matrix into a low-rank matrix and a sparse matrix. In \cite{rpcaJi, video1}, RPCA demonstrates its successful application in surveillance video background modeling, by separating the low-rank stable background and the sparse foreground variations. This model can also be used to provide recognized facial images under variable lighting and pose \cite{ddwPeng, face1}. In the area of recommender systems, RPCA provides recommendations based on the user's preferences, since the data matrix of all user-ratings may be approximately low-rank \cite{recommend1, recommend2, recommend3}.

The RPCA model involves two key subproblems, \textit{i.e.}, $\ell_0$-norm minimization (sparse subproblem) and rank function minimization (low-rank subproblem), which are usually transformed into their convex approximation, \textit{i.e.}, $\ell_1$-norm minimization and nuclear norm minimization. The $\ell_1$-norm minimization problem can be solved by soft thresholding operator \cite{sto}, and the nuclear norm minimization problem can be solved by singular value soft thresholding operator \cite{rpcaJi}. However, these convex approximations usually suffer from the undesirable overshrink problem, which shrinks different components equally. Therefore, Cand\'{e}s \textit{et al.} propose a reweighted $\ell_1$-norm iterative algorithm that assigns the weights to resemble the penalization of $\ell_0$-norm minimization problems, and the weights used for the next iteration are computed from the value of the current solution, showing remarkable performance in the areas of sparse signal recovery, statistical estimation, error correction and image processing \cite{wl1mCandes}. The weighted nuclear norm minimization (WNNM) proposed in \cite{wnnmGu} assigns different weights in a non-descending order, according to the different importance of rank components. Peng \textit{et al.} propose the dual weighted model as a feasible extension way to improve the performance of RPCA, which integrates the weighted low-rank minimization problem and the weighted sparse minimization problem into an iterative method \cite{ddwPeng}. Moreover, based on a high-dimensional geometrical analysis (Gaussian angle analysis) of the null-space characterization, the rigorous theoretical analysis for these weighted methods is provided, which demonstrates the weighted $\ell_1$-norm minimization can indeed deliver recoverable sparsity thresholds, and recover the observations with more serious noise interference \cite{weightsXu}. Recently, the Schatten $p$-norm ($0<p<1$) attracts the attention of many researchers because of its weakly restricted isometry properties \cite{lp1, lp2, lp3, lp4}. In \cite{wsnmXie}, the weighted Schatten $p$-norm minimization (WSNM) is proposed to provide a good performance in video background subtraction problem. However, the squared Frobenius norm fidelity term of WSNM is best suited for the additive Gaussian noise, which cannot obtain a satisfactory performance when the measurements contain gross errors or impluse noise. Thus, the WSNM-RPCA model \cite{wsnmrpcaXie} introduces a sparse constraint penalty based on the $\ell_1$-norm while the WSNM-$\ell_1$ model \cite{Chen} adopt the $\ell_1$-norm as its data fidelity term. Both of these method utilizes the $\ell_1$-norm to capture the sparse structure of the observation data, which further improve the performance in sparse denoising problems. By exploiting the nonlocal self-similarity of natural images, Zhao \textit{et al.} propose a nonconvex model for compressed sensing recovery problems by taking use of the Schatten $p$-norm \cite{reviewer3_2}. In \cite{reviewer3_1}, Zhang \textit{et al.} propose a robust alternating low-rank representation model formed by an alternating forward-backward representation process. It recovers the low-rank components by the RPCA in the forward representation, and then performs a joint $\ell_p$-norm and $\ell_{2,p}$-norm minimization ($0<p<1$) based on the low-rank representation by taking the low-rank components as inputs and dictionary for subspace recovery.The nonconvex $\ell_p$ quasi-norm approaches present a tradeoff: closer approximation of sparsity for harder analysis and computation \cite{Chartrand2}. For this problem, an iterated generalized shrinkage thresholding (GST) algorithm is proposed by iteratively searching for the nonzero solution of $\ell_p$ quasi-norm\cite{gst}, which has been widely used in image denoising application\cite{wsnmXie,wsnmrpcaXie,Chen}. And in \cite{Chartrand3} and \cite{Chartrand1}, Chartrand proposes a generalized $p$-shrinkage operation as the approximate closed-form solution of the $\ell_p$-norm minimization, which has a good global convergence and stability in the nonconvex optimization problem.

The presence of image noise often results in severe image degradation and make the post-processing more challenging. The purpose of image denoising is to remove the noise and recover the clean image from the noisy observation. The salt-and-pepper noise, also known as shot noise or spike noise, is one of the most typical impulse noise that usually introduced by dead pixels in sensor cell malfunction, faulty memory locations in hardware, digital converter errors, or long distance transmission \textit{etc} \cite{rpcaCandes,peppersalt1}. This noise appears as black dots in the bright part like 'pepper' and white dots in the dark part like 'salt' that randomly scattered in an image. The pepper noise is valued as the minimum value and the salt noise is valued as the maximum value in the image. In a salt-and-pepper noise image with the probability $P$ ($0<P \leqslant 1$), $P\%$ pixels are contaminated by the salt noise or pepper noise while the rest $1-P\%$ pixels remain as they are. Then, how to remove the noised pixels without affecting the clean pixels is one of the most challenging problems. The nonlinear filtering technique is widely used in salt-and-pepper denoising due to their good performances and low computational complexity, such as the median filter \cite{med}, the weighted median filter (WMF) \cite{wmf} and the center weighted median filter (CWMF) \cite{cwmf1,cwmf2}. However, these filters process every pixel without checking whether it is an impulse noise or not, leading to the image distortion and the loss of details and edges. The adaptive median filter (AMF) \cite{amf1,amf2} can handle the salt-and-pepper noise with higher probability by adaptively increasing the filtering window size during filter operation. Further, some improved decision-based methods are developed by detecting the corrupted pixels and then performing image restoration with appropriate filtering approaches \cite{dba,idba,pdba}. The classical variational regularization imposes consistency with the image smoothing prior and typically penalizes oscillatory behavior of the noise \cite{variational}. The total variation (TV)-based image regularization \cite{rof} which can be applied to the discontinuous function, enabling the preservation of sharp edges, but it is easily to produce the staircase effect. Some developed TV models are proposed to overcome this problem, such as the four-directional TV \cite{4_TV} and the overlapping group sparsity TV \cite{ogstv1,ogstv2}. In \cite{tvsolution}, a splitting algorithm for bound-constrained TV is proposed to efficiently solve these nondifferentiability total variation functions, which can be extended to other TV-based problems flexibly. However, these classical methods may tend to oversmooth image details, and generate artificial textures when images have high-density noise.

In recent years, the patch-based image restoration schemes that exploit the image nonlocal self-similarity (NSS), has emerged as one promising approach for various image restoration tasks. The block-matching and 3-dimensional filtering (BM3D) algorithm searches for patches which are similar to the currently processed one and stacks the matched patches into a 3D array, then denoises the observation data by the collaborative filter \cite{bm3d1,bm3d2}. In \cite{lr1,lr2,lr3}, the main idea of the low-rank prior-based model is to recover the underlying low-rank structure of a matrix from its degraded/corrupted observation, by stacking the nonlocal similar patch vector into a matrix, which should be low-rank.

In this paper, we focus on the RPCA-based model to remove the sparse noise from the observation data based on the sparsity prior of the salt-and-pepper noise and the nonlocal self-similarity of the clean image. The image matrix has low rank due to the highly correlated pixels, and the noise matrix is sparse in which few nonzero elements are occupied by isolated and randomly distributed noise pixels. Therefore, the RPCA model can be applied to reconstruct the clean image from the noisy observation. The contributions of this paper are summarized as follows: (i) we integrate the $\ell_p$ quasi-norm ($ 0 < p < 1 $) into the weighted method and propose a generalized RPCA-based model, called the dual weighted $\ell_p$-norm minimization (DWLP) model, which combined the weighted Schatten $p$-norm of the low-rank matrix and the weighted $\ell_p$ quasi-norm of the sparsity matrix, enhancing the sparsity and the low-rank simultaneously for matrix recovery; (ii) An approximate closed-form solution is introduced to solve the $\ell_p$-norm minimization, which has more stability in the nonconvex optimization and provides a more accurate estimation for the low-rank and sparse matrix recovery problem; and (iii) we apply the DWLP model to remove the salt-and-pepper noise and provide extensive experimental results to demonstrate its state-of-the-art performance in view of both the quantitative evaluation and the subjective visual quality.

The reminder of this paper is organized as follows: Section~\ref{2} reviews the related work on low-rank and sparse matrix recovery problems and the evolutions of the RPCA model. Section~\ref{3} presents the dual weighted $\ell_p$-norm (DWLP) minimization model and its solution which combined with the inexact augmented Lagrange multiplier (IALM) and the $p$-shrinkage operator to solve the nonconvex optimization effectively. Section~\ref{4} describes the application of the proposed DWLP model to salt-and-pepper denoising based on the nonlocal similar property in nature images, and validates its improvements of DWLP over various variational RPCA-based models. Section~\ref{5} demonstrates the superiority of the proposed DWLP model in the salt-and-pepper denoising problem based on the nonlocal scheme. Then, we provide the parameter setting of the DWLP model for the salt-and-pepper denoising. Finally, we analyze the influence of the DWLP parameter values under different noise levels. In the end, concluding remarks are given in Section~\ref{6}.

\section{Related Work}
\label{2}

\subsection{Sparsity Recovery}
\label{2.1}
Sparsity as one of the most common prior knowledges widely exists in observations, the number of zero-valued elements divided by the total number of elements is called the sparsity of a matrix. For an observation data $\mathbf Y \in \mathbb R ^{m \times n}$, the sparse recovery problem is to find a sparse matrix $\mathbf X$. The sparse matrix $\mathbf X$ is close to $\mathbf Y$ under the ${\mathrm F}$-norm fidelity term which can be enforced by the $\ell_0$-norm regularization term. The sparsity recovery problem based on $\ell_0$-norm minimization model is defined as follows:
\begin{equation}
\hat{ \mathbf X }=\arg \min \limits_{ \mathbf X } \Vert \mathbf X - \mathbf Y \Vert^2_\mathrm{F} +\beta \Vert \mathbf X \Vert_0
\label{eq:1}
\end{equation}
where $\beta>0 $ is the trade-off parameter to balance the data fidelity induced by $\Vert . \Vert^2_\mathrm{F}$ and the sparse regularization term by $\Vert . \Vert_0 $. The $\ell_0$-norm counts the number of nonzero entries of the matrix. Regardless of the entry value, each nonzero entry makes the same contribution to the objective function, indicating the "democratic penalization".

\subsubsection{$\ell_1$-Norm Minimization}
\label{sub:2.1.1}
Although the $\ell_0$-norm minimization model obtains an accurate description for sparse structure recovery, it is a highly nonconvex problem and requires an intractable combinatorial search for solution \cite{ddwPeng}. Therefore, the $\ell_1$-norm as one of the tightest tractable relaxation of the $\ell_0$-norm \cite{weightsXu}, is adopted to formulate the $\ell_1$-norm minimization problem as follows:
\begin{equation}
\hat{ \mathbf X }=\arg\min \limits_{ \mathbf X }\Vert \mathbf X - \mathbf Y \Vert^2_\mathrm{F}+\beta \Vert \mathbf X \Vert_1
\label{eq:2}
\end{equation}
where $\beta>0 $ is the trade-off parameter. It is widely used to recover sparse structures and appears in many sparse prior-based optimization problems as a subproblem (e.g., robust principal component analysis (RPCA) \cite{rpcaJohn,rpcaCandes}, sparse coding \cite{sparseCoding}). In general, the optimum solution of Eq.~\eqref{eq:2} can be obtained through the soft thresholding operation \cite{sto}, \textit{i.e.}, $\hat{x_i}=\mathrm{soft}(y_i, \beta)=\mathrm{sgn}(y_i) \max(\vert y_i \vert -\beta,0)$, where $\hat x_i $ represents the recovered sparsest result of the $i$th sample $y_i $. The soft thresholding operation tend to shrink each larger entries $\vert y_i \vert > \beta $ with the same value $\beta$. This leads to a key difference between the  $\ell_1$-norm and $\ell_0$-norm: unlike the democratic $\ell_0$-norm, the larger entries makes more contribution to the objective function, and penalized more heavily in the $\ell_1$-norm than in the $\ell_0$-norm, indicating the "overshrink problem".

\subsubsection{Weighted $\ell_1$-Norm Minimization}
\label{sub:2.1.2}
The $\ell_1$-norm minimization is not able to provide a perfect solution, for the convex relaxation problem is not equivalent to the original $\ell_0$-norm problem \cite{Zha}. In fact their solutions are supposed to be equal with high probability under some choice of the trade-off parameter $\beta$, instead of exactly the same \cite{Candes}.  whereas the $\ell_1$-norm minimization shrinks each element with the same $\beta$ as indicated in the soft thresholding operation, leading to an inaccurate estimation of the location of nonzero elements in the recovered sparse matrix\cite{ddwPeng}.

To address this imbalance, a more effective regularization norm is proposed to further improve the performance of the $\ell_1$-norm minimization, called the weighted $\ell_1$-norm \cite{ddwPeng, wl1mCandes}, which is defined as $\Vert \mathbf W \odot \mathbf X \Vert_1=\sum\limits^{mn}_{i=1} w_i \vert x_i  \vert$, where $\odot $ is the Hadamard product, and $w_i $ denotes the $i $th weight inversely proportional to the absolute value of $x_i $, \textit{i.e.}, $w_i=\frac{1}{\vert x_i \vert +\epsilon} $, and $ \epsilon $ is a very small constant to avoid the denominator being zero. This regularization term supplies a more reasonable weighting rule to the optimization problem, which enables different entries to make different contributions for the regularization term by discouraging the small but nonzero entries by large weights and encouraging large entries by small weights. Here, the weighted $\ell_1$-norm minimization problem can be defined as follows:
\begin{equation}
\hat{ \mathbf X }=\arg \min \limits_{ \mathbf X } \Vert \mathbf X - \mathbf Y \Vert^2_\mathrm{F} +\beta \Vert \mathbf W \odot \mathbf X \Vert_1
\label{eq:3}
\end{equation}
where $\beta>0 $ is the trade-off parameter.

The weighted $\ell_1$-norm minimization has made an impressively quantitative improvement for recovering sparse signals \cite{wl1mCandes}, since the weights greatly generalize the typical $\ell_1$-norm regularization term that represents a more democratic case similar to the original $\ell_0$-norm.
%Moreover, there is an iterative reweighted algorithm \cite{ddwPeng,wl1mCandes}, which allows to further improve the accuracy of sparse signal recovery, and the 're' means to estimate the elements and the weights alternatively.

\subsection{Low Rank Recovery}
\label{2.2}
For a low-rank matrix $\mathbf X \in \mathbb R ^{m \times n} $, its rank (defined as the number of nonzero singular values) is much less than the number of rows or columns, \textit{i.e.}, $ \mathrm{rank}(\mathbf X )\ll \min(m,n) $, which means the vector of the singular values is sparse. Suppose that $\mathbf Y \in \mathbb R ^{m \times n}$ is the observation data, and $\mathbf X $ denotes its underlying low-rank matrix. The matrix rank minimization problem, which aims to recover the underlying low-rank structure $\hat{ \mathbf X }$ from $\mathbf Y $, is formulated as follows:
\begin{equation}
\hat{ \mathbf X }= \arg \min \limits_{ \mathbf X } \Vert \mathbf X - \mathbf Y \Vert^2_\mathrm{F} +\alpha \, \mathrm{rank} ( \mathbf X )
\label{eq:4}
\end{equation}
where $\alpha>0 $ is the trade-off parameter between the low-rank regularization term and the fidelity term.

The matrix rank minimization problem is designed to capture the intrinsic rank of the observation data exactly, but it is a highly nonconvex and nonlinear problem without efficient solution.

\subsubsection{Nuclear Norm Minimization}
\label{sub:2.2.1}
Since the rank minimization problem cannot be solved directly, the rank function is usually replaced by its tightest convex relaxation---the nuclear norm, also called the trace norm or the Schatten $1$-norm \cite{weightsXu, pcp, sparseCoding}. The nuclear norm minimization (NNM) problem is formulated as follows \cite{rpcaJohn,aboutnnm}:
\begin{equation} \hat{ \mathbf X }= \arg \min \limits_{ \mathbf X } \Vert \mathbf X - \mathbf Y \Vert^2_\mathrm{F} +\alpha \, \Vert  \mathbf X \Vert_*
\label{eq:5}
\end{equation}
where $\alpha>0 $ is the trade-off parameter, the nuclear norm regularization term of matrix $\mathbf X $ is defined as the sum of its singular values, \textit{i.e.}, $\Vert \mathbf X \Vert _* =\sum\limits_{i=1}^r \sigma_i( \mathbf X )$ and $\sigma_i ( \mathbf X ) $ is the $ i $th singular value of $\mathbf X $, obtained by $\mathbf X =\mathbf U \mathbf \Sigma \mathbf V ^T $, $ \mathbf \Sigma = \mathrm{diag} \begin{pmatrix} \sigma_1(\mathbf X), \sigma_2(\mathbf X),\cdots, \sigma_r(\mathbf X) \end{pmatrix} $, $r=\min(m,n) $. Its optimum solution can be effectively solved by the singular value thresholding (SVT) operation \cite{svt} as $\hat{\mathbf X }=\mathbf U~\mathrm{ soft}(\mathbf \Sigma,\alpha )~\mathbf V ^T $, in which $\mathbf \Sigma $ denotes the singular value diagonal matrix of $\mathbf Y =\mathbf U \mathbf \Sigma \mathbf V ^T $, and $\mathrm{soft}(\mathbf \Sigma,\alpha ) $ represents the soft thresholding function of $\mathbf \Sigma  $ with parameter $\alpha $, defined as $\mathrm{soft}(\mathbf \Sigma_{ii},\alpha )=\max(\mathbf \Sigma_{ii}-\alpha,0) $ for each diagonal element $\mathbf \Sigma _{ii} $.
\subsubsection{Weighted Nuclear Norm Minimization }
\label{sub:2.2.2}
Unlike the original nonconvex matrix rank minimization problem with a democratic penalization, the NNM suffers from the overshrink problem since it shrinks different components equally with the same value of $\alpha $ in $\mathrm{soft}(\mathbf \Sigma ,\alpha)$. However, this greatly restricts its capability and flexibility in practice since the singular values have clear physical meanings and should be treated differently.

To overcome the shortcoming of the NNM and better approximate the rank function, the truncated nuclear norm regularization term (TNNR) \cite{tnnr} and the partial sum minimization (PSM) \cite{psm} have been proposed to protect major components by keeping $k$ largest singular values as they are and minimizing the remaining small singular values merely. However, it is hard to estimate the intrinsic rank that determined by the matrix content, and there is only a binary decision to determine whether to regularize a specific singular value or not, let alone to treat each component differently. Thus, both TNNR and PSM are not flexible enough to incorporate the prior knowledge of different singular values. To solve this problem, a more reasonable variant of the NNM problems has been presented by Gu \textit{et al.}, called the weighted nuclear norm minimization (WNNM) \cite{wnnmGu}, which is defined as follows:
\begin{equation}
\hat{ \mathbf X } = \arg \min \limits_{ \mathbf X } \Vert \mathbf X - \mathbf Y \Vert^2_\mathrm{F} +\theta \, \Vert \mathbf X \Vert_{w,*}
\label{eq:6}
\end{equation}
where $\theta>0 $ is the trade-off parameter, $\Vert \mathbf X  \Vert_{w,*}=\sum\limits_{i=1}^r  w_i\sigma_i(\mathbf X ) $ in which $w_i=\frac{1}{\sigma_i(\mathbf{X})+\epsilon} $ denotes the weight of the $i$th singular value $\sigma_i(\mathbf{X}) $. The WNNM assigns different weights to different singular values and guarantees a more accurate recovery. In this way, the larger singular values could be penalized less than the smaller ones, since they are generally associated with major projection orientations, which should be shrunk less to better preserve major data components.

Based on the WNNM, the weighted Schatten $p$-norm (WSNM) \cite{wsnmXie} has been proposed to further improve the performance through the Schatten $p$-norm ($ 0 < p < 1 $) in the low-rank matrix recovery problem. For the observation matrix $\mathbf Y $, the WSNM problem can be formulated as follows:
\begin{equation}
\hat{ \mathbf X } = \arg \min \limits_{ \mathbf X } \Vert \mathbf X - \mathbf Y \Vert^2_\mathrm{F} +\theta \Vert \mathbf X \Vert_{w,\mathrm S_p}^p
\label{eq:7}
\end{equation}
where $\theta>0 $ denotes the trade-off parameter, and the regularization term is defined as $\Vert \mathbf X  \Vert_{w,\mathrm S_p}^p = \sum\limits_{i=1}^r w_i\sigma_i \left(\mathbf X \right)^p  $ with power $ 0 < p < 1 $ and weight $w_i=\frac{1}{\sigma_i(\mathbf X )+\epsilon} $. For these non-ascending singular values, their weights are guaranteed to be non-descending\cite{wnnmGu,wsnmXie}.

The weighted Schatten $p$-norm provides a more feasible scheme to simulate the rank function through shrinking each singular value depending on its magnitude and combing with a weak restriction supported by setting the power $ 0 < p < 1 $, that greatly improves the flexibility in many applications, e.g., image restoration \cite{lr2, imgres}. When the weights satisfy a non-descending order, the optimum solution of WSNM can be achieved by transforming Eq.~\eqref{eq:7} into a series of independent nonconvex $\ell_p$-norm subproblems, which can be solved by the generalized soft thresholding algorithm \cite{gst,wsnmXie,wsnmrpcaXie,gsvt}.

\subsection{Robust Principal Component Analysis}
\label{2.3}
The RPCA \cite{rpcaJohn,rpcaCandes,rpcaJi} model is proposed based on the classical principal component analysis (PCA). The classical PCA model remains optimal in the low-rank matrix recovery problem when the noise is subject to Gaussian distribution, but non-Gaussian noise, even a single gross error, could deviate the estimation far from its ground truth. Thus, the RPCA is modeled as a more robust optimization problem combined with the sparse regularization term, and transforms the recovery problem into a joint low-rank and sparse matrix approximation.

Suppose that a corrupted observation data $\mathbf D \in\mathbb R ^{m\times n} $ is composed of a low-rank matrix $\mathbf A \in\mathbb R ^{m\times n}$ and an error matrix $\mathbf E \in\mathbb R ^{m\times n}$ (assumed to be sparse with arbitrarily large magnitude), \textit{i.e.}, $\mathbf D =\mathbf A +\mathbf E $. The RPCA model is defined as:
\begin{equation}
\min\limits_{\mathbf{A,E}}~\mathrm{rank} (\mathbf A ) + \lambda \Vert \mathbf E \Vert_0\quad \mathrm{s.t.}~ \mathbf A + \mathbf E = \mathbf D
\label{eq:8}
\end{equation}
where parameter $\lambda >0 $ is the trade-off parameter between enforcing the low-rank property and separating the sparse error appropriately. The low-rank regularization term is induced by the rank function, \textit{i.e.}, the number of nonzero singular values in $\mathbf A$, and the sparsity penalty of matrix $\mathbf E$ is induced by $\ell_0 $-norm, \textit{i.e.}, the number of nonzero entries in $\mathbf E$.

Unfortunately, both $ \mathrm{rank}(\mathbf A ) $ and $\Vert \mathbf E \Vert_0 $ are highly nonconvex and nonlinear problems that cannot be solved directly. Thus, under rather weak assumptions, the RPCA problem can usually be approximated by the principal component pursuit (PCP) problem \cite{rpcaCandes, rpcaJi, pcp}, which is defined as follows:
\begin{equation}
\min\limits_{\mathbf{A,E}}~ \Vert \mathbf A  \Vert_* + \lambda \Vert \mathbf E  \Vert_1\quad \mathrm{s.t.} ~\mathbf A + \mathbf E = \mathbf D
\label{eq:9}
\end{equation}
where $\lambda >0 $ is the trade-off parameter, $\Vert \mathbf A \Vert_* $ is the nuclear norm of matrix $\mathbf A $, the tightest convex relaxation of $ \mathrm{rank}(\mathbf A ) $, defined as the sum of its singular values, \textit{i.e.}, $\Vert \mathbf A \Vert_*=\sum\limits_{i=1}^r\sigma_i(\mathbf A ) $, where $\sigma_i(\mathbf A ) $ is the $i $th singular value of $\mathbf A  $; and $\Vert \mathbf E \Vert_1 $ is the $\ell_1 $-norm of matrix  $\mathbf E $ that replaces $\Vert \mathbf E \Vert_0 $ as its tightest convex relaxation, $\Vert \mathbf E \Vert_1=\sum\limits_{i=1}^{mn}\vert e_i \vert $, where $ e_i $ means the $i $th element in $\mathbf E $.

It has been proved that if the PCP satisfies the following conditions: (i) $\lambda=O\left(\frac{1}{\sqrt{t_{(1)}}} \right) $, $t_{(1)}=\max(m,n)$; (ii) $\mathrm{rank}(\mathbf A )\leqslant O\left(\frac{t_{(2)}}{\mathrm{log}(t_{(1)})}\right) $, $t_{(2)}=\min(m,n)$; and (iii) $\Vert \mathbf E  \Vert_0 \leqslant O\left(mn\right) $, it would be able to estimate the unique $\mathbf A $ and $\mathbf E $ exactly from the observation data $\mathbf D $ with a high probability more than $1-O\left(t_{(1)}^{-10}\right) $ \cite{rpcaCandes,rpcaJohn}.

Note that, either the $\ell_1 $-norm minimization problem (convex approximation of the $\ell_0$-norm minimization problem) to recover the sparse matrix, or the nuclear norm minimization problem (convex approximation of the rank function minimization problem) to recover the low-rank matrix, can be regarded as the subproblem included in the process of solving the PCP model and involved to better approximate the RPCA problem. Specifically, taking the WSNM model mentioned in Section~\ref{sub:2.2.2} as the low-rank regularization term, the WSNM-RPCA method could give a better approximation to the original low-rank assumption through the weighted nuclear norm and the weaker power $0<p\leqslant1$, which is defined as follows:
\begin{equation}
\min\limits_{\mathbf{A,E}}~\Vert \mathbf A  \Vert_{w,\mathrm S_p}^p + \lambda \Vert \mathbf E  \Vert_1 \quad \mathrm{s.t.} ~ \mathbf A + \mathbf E = \mathbf D
\label{eq:10}
\end{equation}
where $\lambda>0 $ is the trade-off parameter, and the weight $w_i=\frac{1}{\sigma_i(\mathbf A )+\epsilon} $ in which $\sigma_i(\mathbf A )$ is the $i $th singular value of $\mathbf A $.

In addition to the low-rank regularization term, another important factor that affects the performance of RPCA is the sparse regularization term. Peng \textit{et al.} propose the weighted low-rank matrix recovery scheme to enhance the low-rank constraint penalty and the sparse constraint penalty simultaneously and greatly improve the performance of the low-rank matrix recovery problem \cite{ddwPeng}. Here, the model is defined as follows:
\begin{equation}
\min \limits_{\mathbf{A,E}} \ \Vert \mathbf A \Vert_{\mathbf \Omega ,*}+\lambda\Vert\mathbf W \odot \mathbf E \Vert_1 \quad \mathrm{s.t.} \ \mathbf A +\mathbf E = \mathbf D
\label{eq:11}
\end{equation}
where $\lambda >0 $ is the trade-off parameter, $\Vert \mathbf A \Vert_{\mathbf \Omega,*}=\sum\limits_{i=1}^r \omega_i\sigma_i (\mathbf A)$ denotes the low-rank penalty with diagonal weight matrix $\mathbf \Omega = \mathrm{diag} \begin{pmatrix} \omega_{1},\omega_{2},\cdots,\omega_{r} \end{pmatrix}$, the weight assigned to the $i$th singular value being $\omega_i  =\frac{1}{\sigma_i(\mathbf A)+\epsilon} $, and $\Vert\mathbf W \odot \mathbf E \Vert_1=\sum\limits^{mn}_{i=1} w_i \vert e_i \vert$ is the sparse constraint with the weight of the $i$th element being $w_i=\frac{1}{|e_i|+\epsilon} $.

\section{Dual Weighted $\ell_{p}$-norm Model}
\label{3}
Note that, the sparsity is referred to as few nonzero elements and large number of zero elements. The connection between sparsity and low-rank recovery is that the sparse matrix is reconstructed by regularizing most of the elements to be zero, and the low-rank matrix is reconstructed by regularizing the singular value vector to be sparse\cite{nss2}. The matrix rank function can be represented as the $\ell_0 $-norm of its singular values, \textit{i.e.}, $\mathrm{rank}( \mathbf X ) = \Vert \mathbf \Sigma \Vert_0 $, $\mathbf X =\mathbf U \mathbf \Sigma \mathbf V ^T $. Therefore, the overshrink problem of both the $\ell_1 $-norm minimization and the nuclear norm minimization can be attributed to the difference between the $\ell_1 $-norm and the $\ell_0$-norm, and filling the gap between them would help us to better solve the overshrink problem in the approximation.

\subsection{Weighted $\ell_p$-Norm Minimization}
\label{3.1}
The $\ell_p$ quasi-norm is perceived as the general case of $\ell_1 $-norm with the power $p $ valued in the range of $(0,1] $, which provides a better performance owing to its weaker restricted isometry property for approximating the $\ell_0 $-norm \cite{wsnmXie,wsnmrpcaXie,imgres}. In the $\ell_p$ spaces, the $\ell_1$-norm is the tightest convex relaxation of $\ell_0$-norm, but the nonconvex $\ell_p$ quasi-norm ($0<p<1$) is more similar to the $\ell_0$-norm geometrically, and as $p~(0<p<1)$ get smaller, the large elements get less shrinkage in $\ell_p$ quasi-norm \cite{Chartrand2}. Therefore, we generalize the weighted $\ell_1 $-norm minimization described in Section~\ref{sub:2.1.2} as the weighted $\ell_p$-norm minimization formulated in Eq.~\eqref{eq:12}, which provides a feasible solution to fill the gap between the $\ell_0 $-norm and the $\ell_1 $-norm,
\begin{equation}
\hat{ \mathbf X }=\arg \min \limits_{ \mathbf X } \Vert \mathbf X - \mathbf Y \Vert^2_\mathrm{F} + \varphi \Vert \mathbf W \odot \mathbf X \Vert_q^q
\label{eq:12}
\end{equation}
where $\varphi>0 $ is the trade-off parameter. The sparse regularization term is defined as $\Vert \mathbf W \odot \mathbf X \Vert_q^q=\sum\limits^{mn}_{i=1} w_i\vert x_i  \vert^q$, in which the weight for the $i$th element is $w_i=\frac{1}{\vert x_i \vert+\epsilon} $, with power $0 < q < 1 $ and small constant $ \epsilon $ to avoid a zero denominator.

The widely used $\ell_1 $-norm can be represented as a special case of the weighted $\ell_p$ quasi-norm by setting both the weights  $w_i $ and the power $q$ to 1, \textit{i.e.}, $\Vert \mathbf W \odot \mathbf X \Vert_q^q=\sum\limits^{mn}_{i=1} w_i\vert x_i\vert ^q=\sum\limits^{mn}_{i=1} \vert x_i  \vert=\Vert \mathbf X \Vert_1$ with $w_i=1 $ and $q=1 $.

\subsection{Our DWLP Model}
\label{3.2}
Here, we incorporate the weighted Schatten $p$-norm of the low-rank matrix and the weighted $\ell_p$ quasi-norm model of the sparse matrix into the RPCA problem to formulate the following DWLP optimization,
\begin{equation}
\min \limits_{\mathbf{A,E}} \ \Vert \mathbf A \Vert_{\mathbf \Omega ,S_{p}}^{p}+\lambda\Vert\mathbf W \odot \mathbf E \Vert_{q}^{q}\quad\mathrm{s.t.} ~\mathbf A +\mathbf E= \mathbf D
\label{eq:13}
\end{equation}
where $\lambda >0 $ is the trade-off parameter, $\Vert \mathbf A \Vert_{\mathbf \Omega,\mathrm S_p}^p=\sum\limits_{i=1}^r \omega_i\sigma_i (\mathbf A) ^p$ denotes the low-rank penalty with diagonal weight matrix $\mathbf \Omega = \mathrm{diag} \begin{pmatrix} \omega_{1},\omega_{2},\cdots,\omega_{r} \end{pmatrix}$, and the weight assigned to the $i$th singular value being $\omega_i  =\frac{1}{\sigma_i({\mathbf A})+\epsilon} $. $\Vert\mathbf W \odot \mathbf E \Vert_q^q=\sum\limits^{mn}_{i=1} w_i \vert e_i \vert ^q$ is the sparse constraint, in which the weight of the $i$th element is $ w_i=\frac{1}{|e_i|+\epsilon} $ with power $0<q\leqslant1$.

Combined with the weighted method and the $\ell_p$ quasi-norm, the proposed model provides better approximations to the rank function and the $\ell_0$-norm which effectively enhances the sparsity of $\mathbf E$ estimation and the low-rank of $\mathbf A$ estimation simultaneously for the matrix recovery.

It is noted that the proposed model can greatly generalize the previous work, since all the PCP, WNNM-RPCA, WSNM-RPCA and DWLP($p$=$q$=$1$) are the special cases of the DWLP:
\begin{itemize}
  \item [1)]
    The principal component pursuit (PCP) problem is the tightest convex relaxation of RPCA. The DWLP can reduce to the PCP when the weights $\omega_i=w_i=1$, and the powers $p=q=1$.
  \item [2)]
    The weighted nuclear norm minimization (WNNM) and its RPCA-based provide a better approximation to the rank function by the weighting scheme. The DWLP can turn out to the WNNM-RPCA when the weights $w_i=1$ and the powers $p=q=1$.
  \item [3)]
    Based on the weighted method, the weighted Schatten $p$-norm (WSNM) and its RPCA-based model further improve the performance by integrating the Schatten $p$-norm ($ 0 < p < 1 $) into the weighted method in the low-rank matrix recovery problem. The DWLP can reduce to the WSNM-RPCA model when the weights $w_i=1$ and the power $q=1$. The WSNM-RPCA is the special case of the proposed model which only enhances the low-rank property.
  \item [4)]
    The DWLP($p$=$q$=$1$) model enhances the sparsity and the low-rank simultaneously by using the weighted method. The DWLP can reduce to the DWLP($p$=$q$=$1$) model when the powers $p=q=1$.
\end{itemize}

To sum up, our DWLP model integrates both the weighted method and $\ell_p$ quasi-norm into the low-rank regularization term and the sparsity regularization term respectively to solve the low-rank and sparse matrix recovery problem, which obtains a significant improvement over the widely used PCP, WNNM-RPCA, WSNM-RPCA and DWLP($p$=$q$=$1$).

\subsection{Solution}
\label{3.3}
Now, let us focus on Eq.~\eqref{eq:13} and invoke the inexact augmented Lagrange multiplier (IALM) algorithm \cite{alm} to approximate its optimum solution alternatively. We first translate Eq.~\eqref{eq:13} into its Lagrange form as follows:
\begin{equation}
\begin{split}
       \mathrm{L}(\mathbf{A,E,Z},\mu) =  &\lambda_a \Vert \mathbf A \Vert_{\mathbf \Omega,\mathrm S_p}^p + \lambda_e \Vert \mathbf W \odot \mathbf E \Vert_q^q\\
                                      &+\langle \mathbf Z, \mathbf D -\mathbf A -\mathbf E\rangle+\frac{\mu}{2} \Vert \mathbf D -\mathbf A -\mathbf E \Vert_\mathrm{F}^2
\end{split}
\label{eq:14}
\end{equation}
where $\lambda_a $, $\lambda_e $ denote the nonnegative trade-off parameters, $\langle \cdot~,~\cdot\rangle$ is the inner product, $\mathbf Z $ and $\mu $ represent the Lagrange multiplier and the penalty parameter respectively. Further, although the parameter $\lambda_a $ can be absorbed into $\lambda_e $, the low-rank regularization term and the sparsity regularization term are solved separately during the procedure of the iterative algorithm \cite{wsnmrpcaXie}. Thus, we prefer to maintain these two parameters for more flexibility. Then, we alternatively update the low-rank matrix $\mathbf A$, the sparse matrix $\mathbf E$, the inexact augmented Lagrange multiplier $\mathbf Z$, and $\mu$. The process of solving the Lagrange function of DWLP is summarized in Algorithm~\ref{alg:1}, where $k$ counts the iteration.

\begin{algorithm}%[H]
    \caption{\textbf{IALM for DWLP-RPCA}}
        \begin{algorithmic}[1]
            \Require Observed data $\mathbf D $, low-rank power $p$, sparse power $q$, and trade-off parameters $\lambda_a$,~$\lambda_e$;  %input:
                \State Initialize: $\mu_0>0$,$~\rho>0$,$~k=0$,$~{\mathbf A}={\mathbf D}$,$~ {\mathbf Z}=0 $;
                \While{not converged}

                \State${\mathbf E}^{k+1} = \arg\min\limits_{\mathbf E} \lambda_e\Vert \mathbf W \odot \mathbf E \Vert_q^q$

                $\quad \quad\quad+ \frac{\mu^k}{2} \Vert \mathbf D + {\mathbf Z}^k/{\mu^k}-{\mathbf A}^k-\mathbf E \Vert_\mathrm{F}^2$;

                 \State${\mathbf A}^{k+1} = \arg\min\limits_{\mathbf A} \lambda_a\Vert \mathbf A \Vert_{{\mathbf \Omega}, {\mathrm S}_p}^p$

                 $\quad\quad\quad + \frac{\mu^k}{2} \Vert \mathbf D + {\mathbf Z}^k/{\mu^k}-{\mathbf E}^{k+1}-\mathbf A \Vert_\mathrm{F}^2$;

               %\state${\mathbf A}_{k+1} = \arg\min\limits_{\mathbf A} \lambda_a\Vert \mathbf A \Vert_{ {\mathbf \Omega}, {\mathrm S}_p^p$

               % \[
               %   \begin{split}\mathbf E_{k+1} = &\arg\min\limits_{\mathbf E} %\lambda_e\Vert \mathbf W \odot \mathbf E \Vert_q^q  \\
               %    &+ \frac{\mu_k}{2} \Vert \mathbf D + \mu_k^{-1} {\mathbf  %Z}_k-{\mathbf A}_k-\mathbf E \Vert_\mathrm{F}^2
               %   \end{split}
                %   \]

                   %\State
                   %\begin{equation*}
                   % \begin{split}
                   %  \mathbf A_{k+1} = &\arg\min\limits_{\mathbf A} %\lambda_a\Vert \mathbf A \Vert_{ \mathbf \Omega, \mathrm S_p %}^p \\
                   % & + \frac{\mu_k}{2} \Vert \mathbf D + \mu_k^{-1}{\mathbf %Z}_k-{\mathbf E}_{k+1}-\mathbf A \Vert_\mathrm{F}^2
                   % \end{split}
                   %\end{equation*}

                      \State  $ \mathbf Z^{k+1} = \mathbf Z^k+\mu^k(\mathbf{D}-\mathbf{A}^{k+1}-\mathbf{E}^{k+1})$;
		              \State  $ \mu^{k+1}=\rho \times \mu^k $;
                      \State  $ k=k+1$;
               \EndWhile
            \Ensure $\hat{ \mathbf A }=\mathbf A^{k+1}$ and $\hat{ \mathbf E }=\mathbf E^{k+1}$ %output:
        \end{algorithmic}
        \label{alg:1}
\end{algorithm}

The minimization of Eq.~\eqref{eq:14} involves two minimization subproblems, \textit{i.e.}, $\mathbf E$ and $\mathbf A$ subproblem. Next, we will present the solutions to the subproblems of $\mathbf E$ and $\mathbf A$ as follows:

1) $\mathbf E$ subproblem: Given the weight matrix $\mathbf W$, the power $q$ and the trade-off parameter $\lambda_e$, the $\mathbf E$ subproblem for each observation data $\mathbf D$ is,
\begin{equation}
\mathbf E^{k+1} = \arg\min\limits_{\mathbf E} \lambda_e\Vert \mathbf W \odot \mathbf E \Vert_q^q + \frac{\mu^k}{2} \Vert \mathbf D + {\mathbf Z}^k/{\mu^k}-{\mathbf A}^k-{\mathbf E} \Vert_\mathrm{F}^2
\label{eq:15}
\end{equation}
this is the Lagrange function of the weighted $\ell_p$-norm minimization stated in Section~\ref{3.1}, which aims to estimate the sparsest solution of DWLP that can be simplified as the following equation,
\begin{equation}
\mathbf E^{k+1} = \arg\min\limits_{\mathbf E} \frac{1}{2} \Vert {\mathbf E}-{\mathbf F}\Vert_\mathrm{F}^2 + \Vert \mathbf \Phi \odot \mathbf E \Vert_q^q
\label{eq:16}
\end{equation}
where ${\mathbf F}={\mathbf D} + {{\mathbf Z}^k} / {\mu^k} - {\mathbf A^k}$ and ${\mathbf \Phi}=\lambda_e{\mathbf W} / {\mu^k} $.

Unfortunately, the Eq.~\eqref{eq:16} cannot be solved by the traditional soft thresholding operation, since the added weights destroy the convexity, and the included $\ell_p$ quasi-norm makes the nonconvex relaxation problem much more difficult to be optimized. Thus, we decompose the problem into $m \times n $ independent $\ell_p$-norm subproblems to reduce the challenge and obtain the global optimum, the $i$th subproblem is given as follows:
\begin{equation}
\hat e_i =\arg \min \limits_{e_i} \frac{1}{2} (e_i-f_i)^2+\phi_i \vert e_i \vert^q
\label{eq:17}
\end{equation}
where $e_i $ denotes the $i$th element in $\mathbf E^{k+1}$, $f_i$ is the $i$th element in $\mathbf F$, \textit{i.e.}, $f_i=d_i+{z_i} / {\mu}-a_i$ and $\phi_i$ is the $i$th element in $\mathbf \Phi$,  \textit{i.e.}, $\phi_i={\lambda_e}{w_i} / {\mu}$. To avoid confusion, the subscript $k$ of $\mu^k$ is omitted for conciseness.

It is noted that the Eq.~\eqref{eq:17} is a nonconvex $\ell_p$-norm minimization, which often solved by the generalized soft thresholding algorithm\cite{gst, wsnmrpcaXie, wsnmXie, Zhang, Zha2}. Here, we adopt the $p$-shrinkage operator \cite{Chartrand1} as the approximate closed-form solution of the $\ell_p$-norm minimization, which has more stability in the nonconvex optimization problem and provides a more accurate estimation for the low-rank and sparse matrix recovery problem.

Therefore, the ${\mathbf E}$ subproblem in the IALM algorithm can be approximately solved through the $p$-shrinkage operator combined with the weighted method by processing each element as:
\begin{equation}
{\hat e}_i^{k+1} =\mathrm{shrink}_p(f_i, \phi_i^k)
\label{eq:18}
\end{equation}
where $\hat{e}_i^{k+1}$ represents the recovered sparsest result of the $i$th sample $f_i$ in the $k$+$1$th iteration, and the $\phi_i^k={\lambda_e}{w_i^k} / {\mu}$ with $w_i^k=\frac{1}{|e_i^{k}|+\epsilon} $ in which $e_i^{k}$ is calculated from the $k$th iteration. The complete $p$-shrinkage operator combined with the weighted method is summarized in Algorithm ~\ref{alg:2}.

\begin{algorithm}[H]
    \caption{\textbf{The $p$-Shrinkage Algorithm Combined with The Weighted Method}}
        \begin{algorithmic}[1]
            \Require data ${\mathbf F}$, parameter ${\mathbf \Phi}_i$, power $q$; %$\mathbf F$ %input:
              \State Initialize: $\phi_i=1$;
	
              \For{each $f_i$ in $\mathbf F$}

                \State $\phi_i^k={\lambda_e}{w_i^k} / {\mu}$ with $w_i^k=\frac{1}{|e_i^{k}|+\epsilon} $;

                \State $\hat{e}_i^{k+1} =\mathrm{sgn}(f_i) \mathrm{max}(0, \vert f_i \vert - \phi_i^k\vert f_i \vert^{q-1})$;

              \EndFor

		\Ensure $\hat{\mathbf E}$

        \end{algorithmic}
        \label{alg:2}
\end{algorithm}

2) $\mathbf A$ subproblem: Given the weight matrix $\mathbf \Omega$, the power $p$ and the parameter $\lambda_a$, the $\mathbf A$ subproblem for each observation data $\mathbf D$ can be written as,
\begin{equation}
\mathbf A^{k+1} = \arg\min\limits_{\mathbf A} \lambda_a\Vert \mathbf A \Vert_{ \mathbf \Omega, \mathrm S_p }^p + \frac{\mu}{2} \Vert \mathbf D + {{\mathbf Z}^k}/{\mu}-{\mathbf E}^{k+1}-\mathbf A \Vert_\mathrm{F}^2
\label{eq:19}
\end{equation}
this is the Lagrange function of the weighted Schatten $p$-norm minimization stated in Section~\ref{sub:2.2.2}. In the non-descending weight order, the WSNM can be effectively solved by the following theorem:

{\bfseries Theorem 1} \cite{wsnmXie}: Let $\mathbf D + \mathbf Z^k / \mu-\mathbf E^{k+1} = \mathbf U \mathbf \Sigma {\mathbf V}^\mathrm T$, $\mathbf \Sigma = \mathrm{diag} \begin{pmatrix} \sigma_1,\sigma_2,\cdots,\sigma_r \end{pmatrix}$. Suppose that all singular values are in non-ascending order and all the weights satisfy $0 \leqslant \omega_1 \leqslant \omega_2 \leqslant \cdots \leqslant \omega_r$, the optimum solution of Eq.~\eqref{eq:18} will be $\hat {\mathbf A}=\mathbf U \mathbf \Delta \mathbf V ^\mathrm T $, with $ \mathbf \Delta =\mathrm{diag}\begin{pmatrix} \delta_1, \delta_2, \cdots, \delta_r\end{pmatrix}$.

    The $i$th singular value $\delta_i$ in $\mathbf \Delta$ can be obtained by solving the independent subproblem in the following,

\begin{equation}
\hat \delta_i =\mathrm{arg} \min \limits_{ \delta_i } \frac{1}{2}(\delta_i-\sigma_i)^2 + \psi_i\delta_i^p
\label{eq:20}
\end{equation}
where ${\psi_i}$ is the $i$th element in ${\mathbf \Psi}$ in which ${\mathbf \Psi}= \lambda_a {\mathbf \Omega} / \mu$ and ${\psi_i}= \lambda_a \omega_i / \mu$.

The $\mathbf A$ subproblem in the IALM algorithm can be approximately solved through the p-shrinkage operator combined with the weighted method by processing each singular value as:
\begin{equation}
\delta_i^{k+1} =\mathrm{shrink}_p(\sigma_i, \psi_i^k)
\label{eq:21}
\end{equation}
where $\delta_i^{k+1}$ is the recovered singular value in the $k$+$1$th iteration. The $p$-shrinkage operator $\mathrm{shrink}_p(\sigma_i, \psi_i^k) = \mathrm{sgn}(\sigma_i) \mathrm{max}(0, \vert \sigma_i \vert - \psi_i^k\vert \sigma_i \vert^{p-1})$ in which ${\psi_i^k}= \lambda_a \omega_i^k / \mu$ with power $\omega_i  =\frac{1}{\delta_i^k+\epsilon} $. Note that each optimum solution $\hat \delta_i$ inside $\mathbf \Delta$ should be arranged in a non-ascending order so as to preserve the effectiveness of this algorithm.

%\begin{equation}
%\delta_i^{(l)} = \mathrm{GST}(\sigma_i,~\lambda_a \mu^{-1}\omega_i^{(l)} ,~p)
%\label{eq:22}
%\end{equation}
%\begin{equation}
%\omega_i^{(l+1)}=\frac{1}{ \delta_i^{(l)} +\epsilon}
%\label{eq:23}
%\end{equation}
%where $l$ counts the reweighting iteration for the low-rank minimization.

%\begin{algorithm}[H]
%    \caption{\textbf{Reweighted Schatten $p$-norm Minimization via IRGA}}
%        \begin{algorithmic}[1]
%           \Require $ \mathbf \Psi = \mathbf D + \mu^{-1}\mathbf Z_k-\mathbf E_{k+1} $, power $p$, parameters $\lambda_a$ and %           $\mu^{-1}$; %input:
%            \State  $ \mathbf \Psi = \mathbf U \mathbf \Sigma {\mathbf V}^\mathrm T, \mathbf \Sigma = \mathrm{diag} \begin{pmatrix} %            \sigma_1,\sigma_2,\cdots,\sigma_r \end{pmatrix} $;
%                \State  $ \mathbf \Delta =\mathrm{IRGA}(\mathbf \Sigma, \lambda_a\mu^{-1}, p) $;
%                \State  $\mathbf A =\mathbf U \mathbf \Delta \mathbf V ^\mathrm T , \mathbf \Delta = \mathrm{diag} \begin{pmatrix} %               \delta_1,\delta_2,\cdots,\delta_r \end{pmatrix} $;
%            \Ensure Matrix $\mathbf A $ %output:
%        \end{algorithmic}
%        \label{alg:4}
%\end{algorithm}
%Finally, terminate on convergence or when IRGA attains a specified maximum number %of iterations.

\section{Nonlocal-based DWLP for salt-and-pepper Denoising}
\label{4}
Many nonlocal-based low-rank models have witnessed tremendous progress in the field of image restoration \cite{bm3d1, Zha, nss2, nss1} in recent years. It is based on the fact that the matrix formed by stacking nonlocal similar patches from a natural image is of low rank. Some salt-and-pepper noise matrix is sparse since few elements are nonzero noise pixels. Consequently, we demonstrate the performance of our model in the salt-and pepper denoising application.
%{\color{gray}[This sentence can be deleted:]Specifically, we select nonlocal similar patches for a local patch and stretch every patches to a patch vector. Then a data matrix with nonlocal self-similarity (NSS) can be constructed by stacking all the similar patches into a matrix column by column. Obviously, the NSS matrix satisfies the low-rank property because all the patches in each data matrix have similar structures.}
\subsection{General Process}
\label{4.1}
In general, the low-rank and sparse matrix recovery problems combined with the nonlocal method for image restoration include three main steps as follows: (i) Match and group patches to obtain the nonlocal self-similarity (NSS) matrix; (ii) Decompose the NSS matrix into one low-rank matrix and one sparse matrix; and (iii) Aggregate these patches into the reconstructed image.

We process the image patch by patch, for a local patch with size $\sqrt c \times\sqrt c$ extracted from the noisy image, we select $K$ of its nonlocal similar patches based on the Euclidean distance, and stretch every patch to a vector ${\bm d}_i\in {\mathbb R}^c$. After that, all the similar patches are stacked into a data matrix column by column to construct the NSS matrix $\mathbf{D}_i=\begin{Bmatrix}{\bm d}_{i,1},{\bm d}_{i,2},\cdots,{\bm d}_{i,K}\end{Bmatrix} \in \mathbb R ^{c\times K}$, where $\bm{d}_{i,j},~j=1, \cdots,K$ denotes the $j$th similar patches of $\bm{d}_{i}$. Obviously, the NSS matrix satisfies the low-rank property since all the patches in each data matrix have similar structures.

Following \cite{rpcaJi} and \cite{prefilterMatching}, we search for similar patches in the prefiltered images obtained by smoothing the noisy image with a median filter, so as to produce a much more accurate patch matching result than directly in corrupted images. This is because the performance of patch matching will seriously degrade in the presence of severe salt-and-pepper noise. Therefore, we extract the similar patches are from the noisy image with the same index location of its similar prefiltered patches. It is noted that the prefiltered image is only used for patch matching, while the noisy image is used as the input of the denoising process.

In our case, the observation data matrix $\mathbf D_i $ can be considered as a sum of a low-rank image matrix $\mathbf A_i$ and a sparse noise matrix $\mathbf E_i$, \textit{i.e.}, ${\mathbf D}_i = {\mathbf A}_i + {\mathbf E}_i  $, which is usually simplified as ${\mathbf D}={\mathbf A}+{\mathbf E} $. The low-rank and sparse matrix approximation could estimate $\mathbf A $ and $\mathbf E$ under the observation $\mathbf D$. The first column of $\mathbf D$ is the patch to be recovered, and the first column of $\mathbf A$ is the recovered result.

Finally, the reconstructed image can be obtained by aggregating all of the recovered patches. In the nonlocal-based method, image patches are always overlapped and most of the pixels are covered by several recovered patches. Thus, each estimated pixel of the final reconstructed image would be generated by taking the average of the recovered patches at each location. Such a kind of operation is referred to as the patch aggregation, which is used to suppress possible artifacts caused by block discontinuities in the neighborhood of the boundaries of patches \cite{wsnmrpcaXie}.

\subsection{Analysis}
\label{4.2}
\begin{figure}[t!]
\centering
\subfigure[]{\includegraphics[width=1.3in]{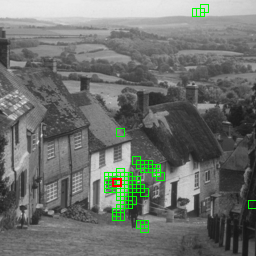}
\label{fig:1 subfig:a}}
\hfil
\subfigure[]{
\includegraphics[width=1.8in]{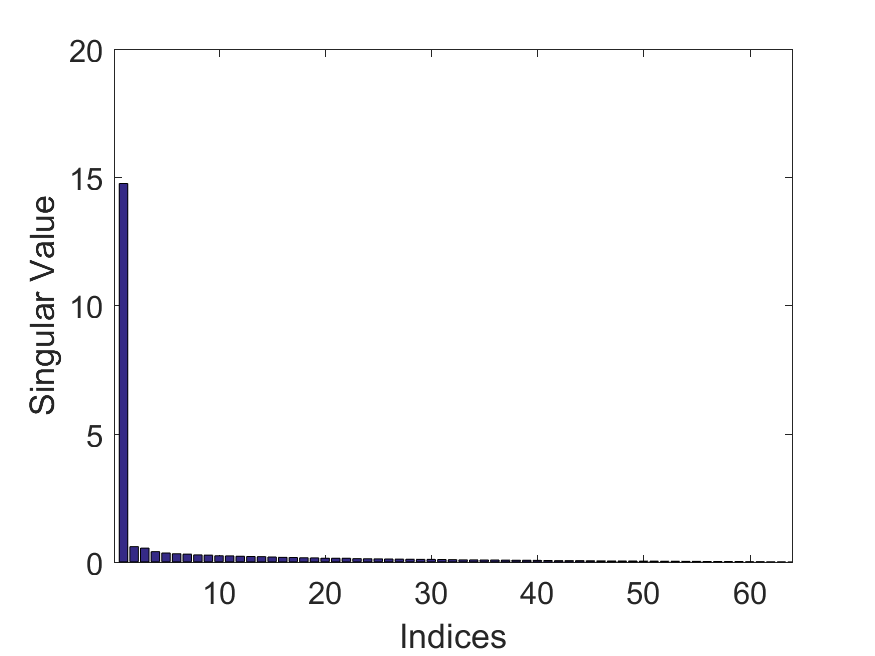}
\label{fig:1 subfig:b}}
\hfil
\subfigure[]{
\includegraphics[width=1.3in]{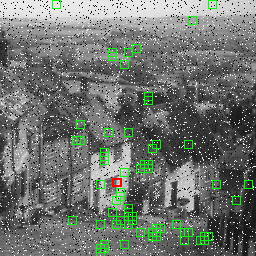}
\label{fig:1 subfig:c}}
\hfil
\subfigure[]{
\includegraphics[width=1.8in]{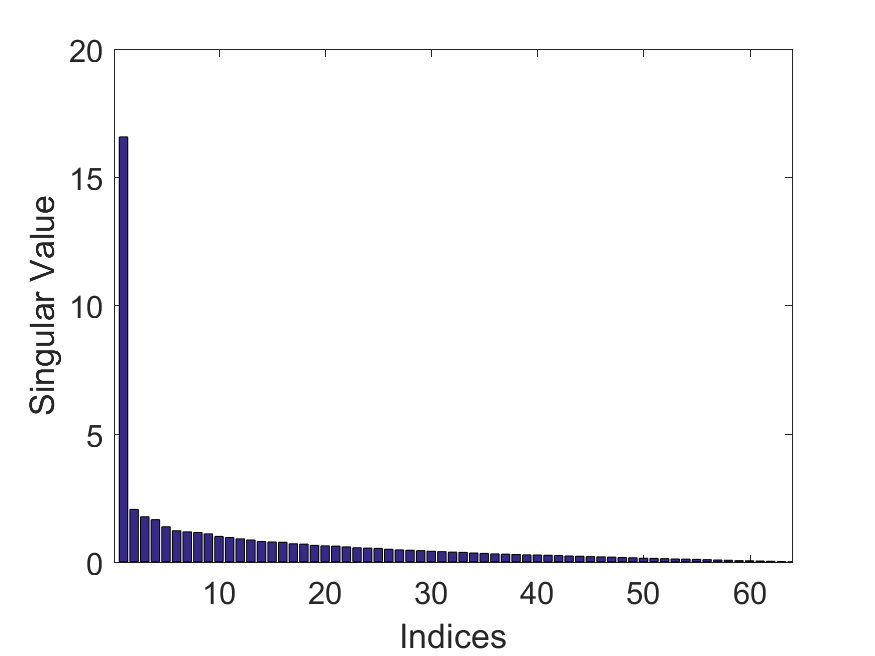}
\label{fig:1 subfig:d}}
\caption{Low rank analysis of a natural image. (a) The test image ({\textit{Goldhill}}), (b) The singular value distribution of clean NSS matrix, (c) Noisy image with 10\% salt-and-pepper noise, (d) The singular value distribution of noisy NSS matrix.  Low rank analysis of a natural image.}
\label{fig:1}
\end{figure}

\begin{figure}[t!]
\centering
\subfigure[]{\includegraphics[width=1.3in]{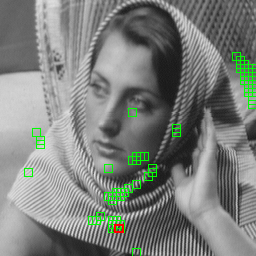}
\label{fig:2 subfig:a}}
\hfil
\subfigure[]{
\includegraphics[width=1.8in]{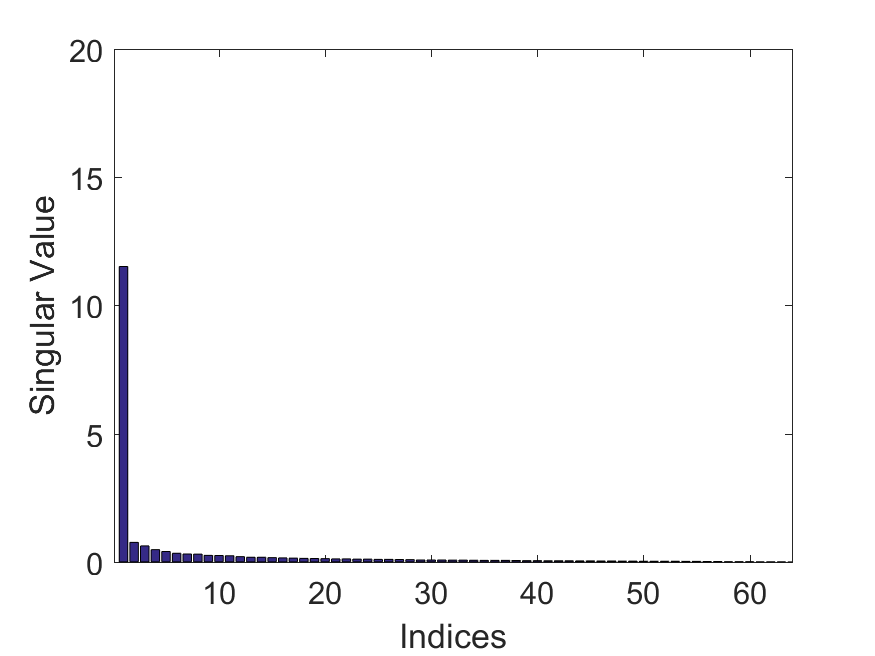}
\label{fig:2 subfig:b}}
\hfil
\subfigure[]{
\includegraphics[width=1.3in]{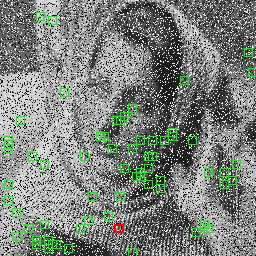}
\label{fig:2 subfig:c}}
\hfil
\subfigure[]{
\includegraphics[width=1.8in]{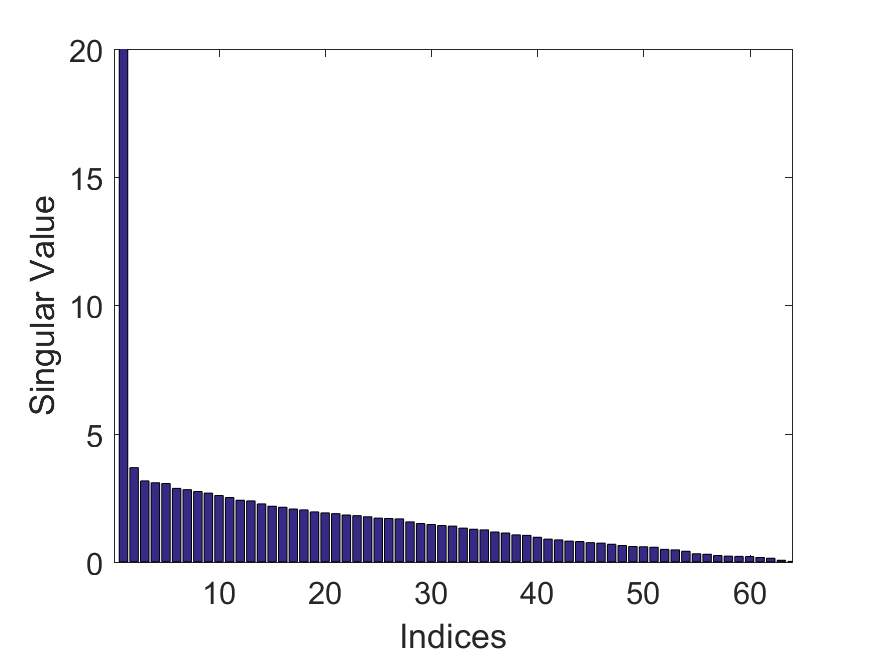}
\label{fig:2 subfig:d}}
\caption{Low rank analysis of a natural image. (a) The test image (\emph{Barbara}); (b) The singular value distribution of clean NSS matrix; (c) Noisy image with 30\% salt-and-pepper noise; (d) The singular value distribution of noisy NSS matrix.}
\label{fig:2}
\end{figure}

We randomly select two NSS matrices and compute their singular values from Fig.~\ref{fig:1 subfig:a} shows a scene of streets, houses, and terraces in the distance, and Fig.~\ref{fig:2 subfig:a} shows a portrait of a woman. One can observe from Figs.~\ref{fig:1 subfig:b} and \ref{fig:2 subfig:b} that the singular values decay very fast, according the low-rank structure of nature images. The NSS matrix can usually be projected into a lower dimensional space owing to the similarity between its column vectors. As shown in Figs.~\ref{fig:1 subfig:c} and \ref{fig:2 subfig:c}, we add 10\% and 30\% salt-and-pepper noise to these two images respectively, where the reference patch is marked by the red box and its similar patches are marked by green boxes. We observed that the NSS matrices extracted from the corrupted images are full rank in Figs.~\ref{fig:1 subfig:d} and \ref{fig:2 subfig:d}. This is due to the fact that the added noise disturbs the similarity between the column vectors in the NSS matrix, and destroys its original low-rank property.
%\Figure[t!](topskip=0pt, botskip=0pt, midskip=0pt){}{Low rank analysis of a natural image}

\xdef\xfigwd{\the\wd\figbox}% <========================================

\xdef\xfigwd{\the\wd\figbox}% <========================================

%\begin{figure*}[!htpb]
%\centering
%\subfigure[ ]{\label{fig:2 subfig:a}
%\includegraphics[width=0.4\linewidth]{NSS_a_Barbara.png}}
%\hspace{0.03\linewidth}
%\subfigure[ ]{\label{fig:2 subfig:b}
%\includegraphics[width=0.5\linewidth]{NSS_b_Barbara.png}}
%\vfill
%\subfigure[ ]{\label{fig:2 subfig:c}
%\includegraphics[width=0.4\linewidth]{NSS_c_Barbara.png}}
%\hspace{0.03\linewidth}
%\subfigure[ ]{\label{fig:2 subfig:d}
%\includegraphics[width=0.5\linewidth]{NSS_d_Barbara.png}}
%\caption{Low rank analysis of a natural image. (a) The test image (\emph{Barbara}); (b) The singular value distribution of clean NSS matrix; (c) Noisy image with 30\% salt-and-pepper noise; (d) The singular value distribution of noisy NSS matrix.}
%\label{fig:2}
%\end{figure*}
Using this low-rank property as the prior knowledge of clean images, there are many image restoration methods to recover the potential low-dimensional subspace of NSS matrices. In order to demonstrate the superiority of the proposed DWLP model to the other nonlocal-based RPCA methods, we conduct a matrix rank singular value decomposition (SVD) experiment, which helps to observe low-rank structures of NSS matrices and illustrates the overshrink problem visually.

\begin{figure}[!htpb]
\centering
\subfigure[ ]{\label{fig:3 subfig:a}
\includegraphics[width=0.7\linewidth]{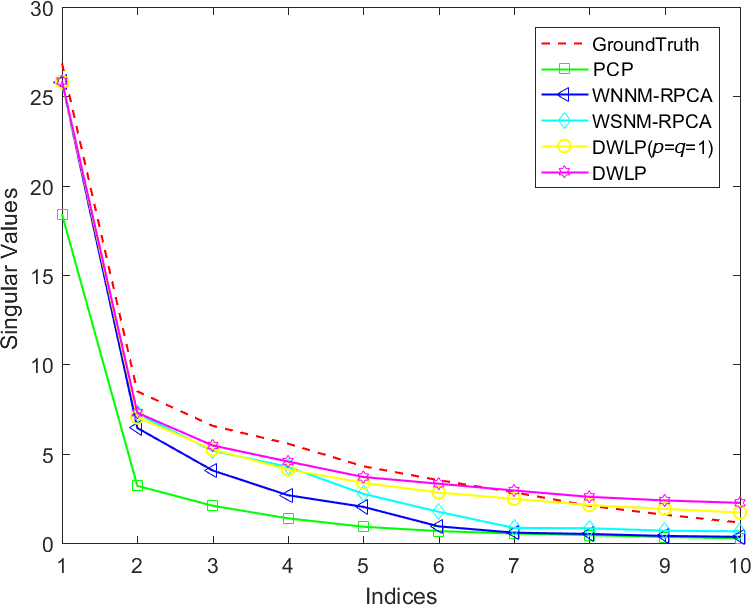}}
%\hspace{0.001\linewidth}
\subfigure[ ]{\label{fig:3 subfig:b}
\includegraphics[width=0.7\linewidth]{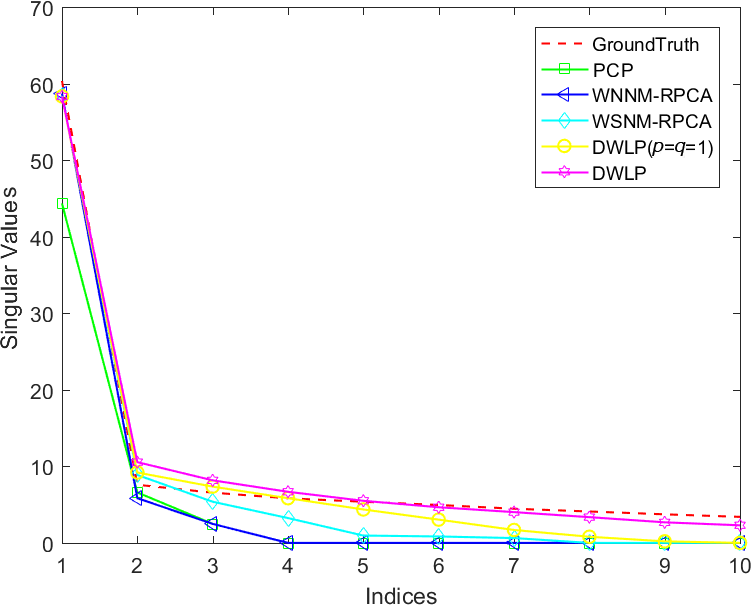}}
\caption{Singular value decomposition (SVD) for NSS matrices. (a) SVD results of NSS matrices sampled from Fig.~\ref{fig:1 subfig:c}; (b) SVD results of NSS matrices sampled from Fig.~\ref{fig:2 subfig:c}.}
\label{fig:3}
\end{figure}

In this experiment, we randomly select 10 NSS matrices, each of which contains 64 similar patches of size 8$\times$8 as shown in Figs.~\ref{fig:1 subfig:c} and \ref{fig:2 subfig:c}, and decompose each NSS matrix into a low-rank matrix and a noise matrix through PCP \cite{rpcaJohn}, WNNM-RPCA \cite{wnnmGu,wsnmrpcaXie}, WSNM-RPCA \cite{wsnmrpcaXie}, DWLP ($p$=$q$=$1$) \cite{ddwPeng}, and DWLP respectively. We observed from their singular value distributions plotted in Fig.~\ref{fig:3} that the classical PCP (denoted by green line) deviates far from the ground truth (denoted by red line), meaning that the overshrinkage is serious. Whereas the large singular values of the WNNM-RPCA method (denoted by blue line) are shrunk less to preserve the major data components with the shrinkage being proportional to the non-descending weights \cite{wnnmGu,wsnmrpcaXie}. When the power $p$ is less than $1$, the nonzero singular values estimated by the WSNM-RPCA method are getting closer to the ground truth (marked by cyan line), indicating the effectiveness of the weak power in recovery methods. However, the overshrink problem not only occours in the low-rank reconstruction but also in the sparse noise estimation, and thus the DWLP($p$=$q$=$1$) achieves a better assumption of the original low-rank structure thanks to its dual weighted scheme (see the yellow line), while the proposed DWLP (denoted by magenta line) further improves the performance by integrating the weights and the $\ell_p$ quasi-norm into both the low-rank and sparse regularization term.

We come to the conclusion through this experiment that our DWLP could better deal with the overshrink problem, and provide a more satisfactory solution of the low-rank and sparse matrix approximation problem.

\section{Experimental Results and Discussions}
\label{5}
In this section, we validate the performance of the proposed DWLP method for image denoising under different salt-and-pepper noise levels and present both qualitative and quantitative comparisons with other state-of-the-art methods. To this end, we analyze the convergence of the proposed DWLP method and the influence of several important paraments. All experiments were run in Matlab R2016a on a personal computer with an Intel Xeon E3-1245 v6 CPU with 3.7GHz and 16GB RAM.

\subsection{Experiments on 10 Test Images}
\label{5.1}
\begin{figure}[!htpb]
\centering
\vspace{.1in}
\includegraphics[width=0.19\linewidth]{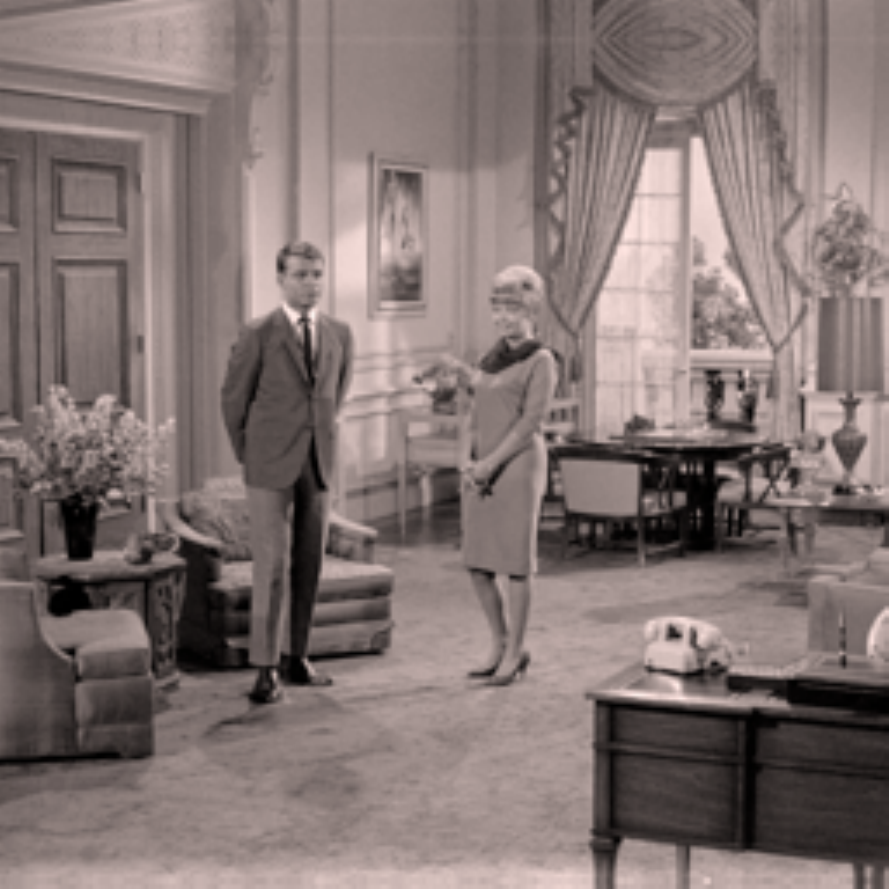}
\includegraphics[width=0.19\linewidth]{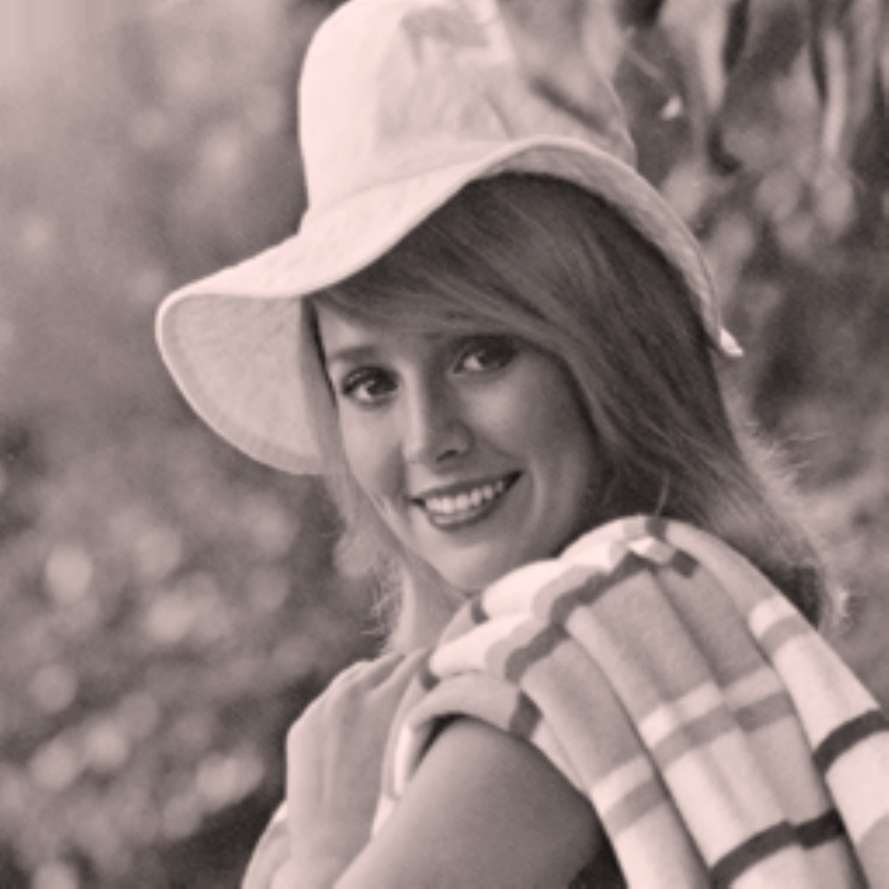}
\includegraphics[width=0.19\linewidth]{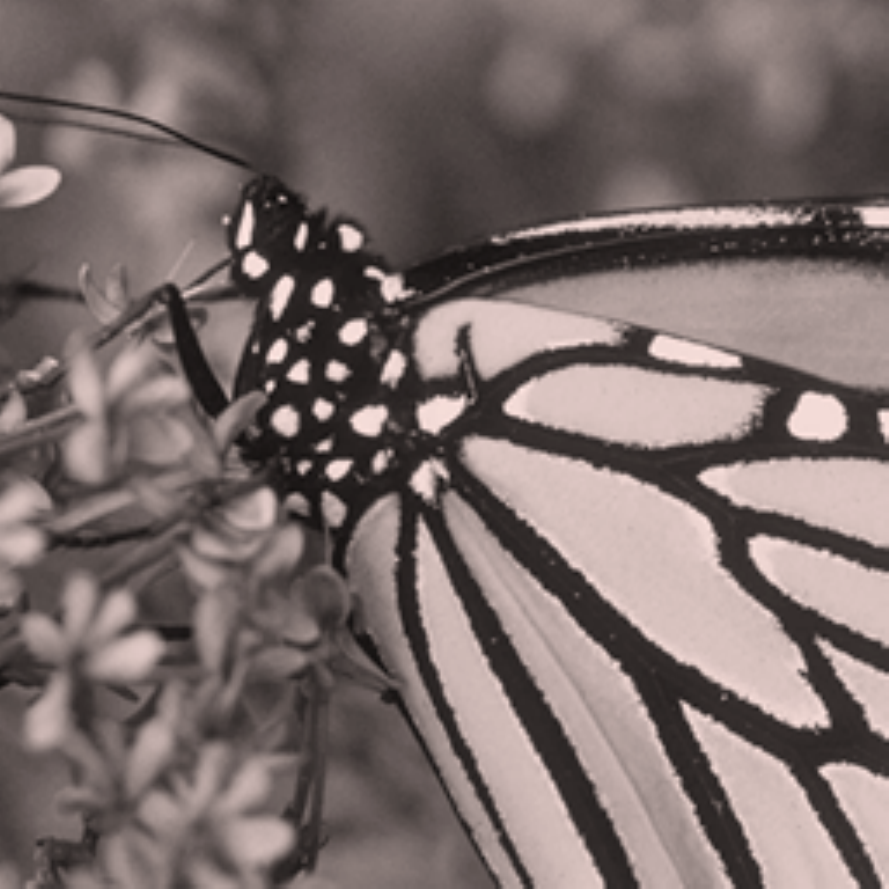}
\includegraphics[width=0.19\linewidth]{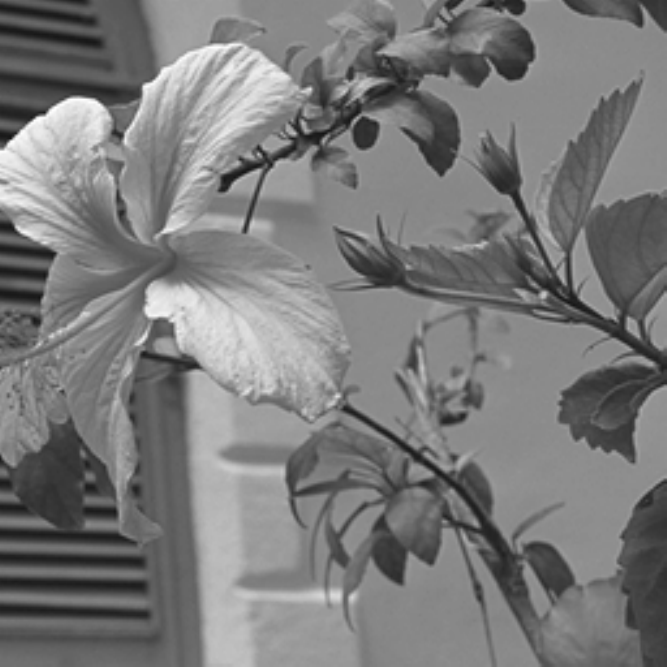}
\includegraphics[width=0.19\linewidth]{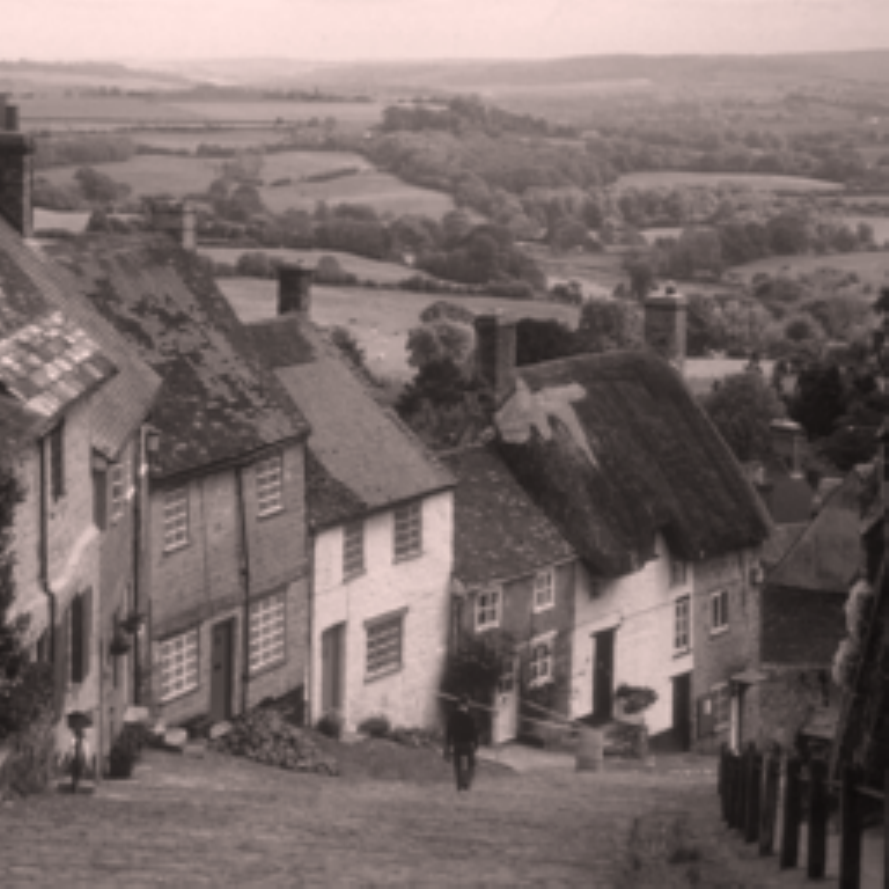}
\vfill
\vspace{.03in}
\includegraphics[width=0.19\linewidth]{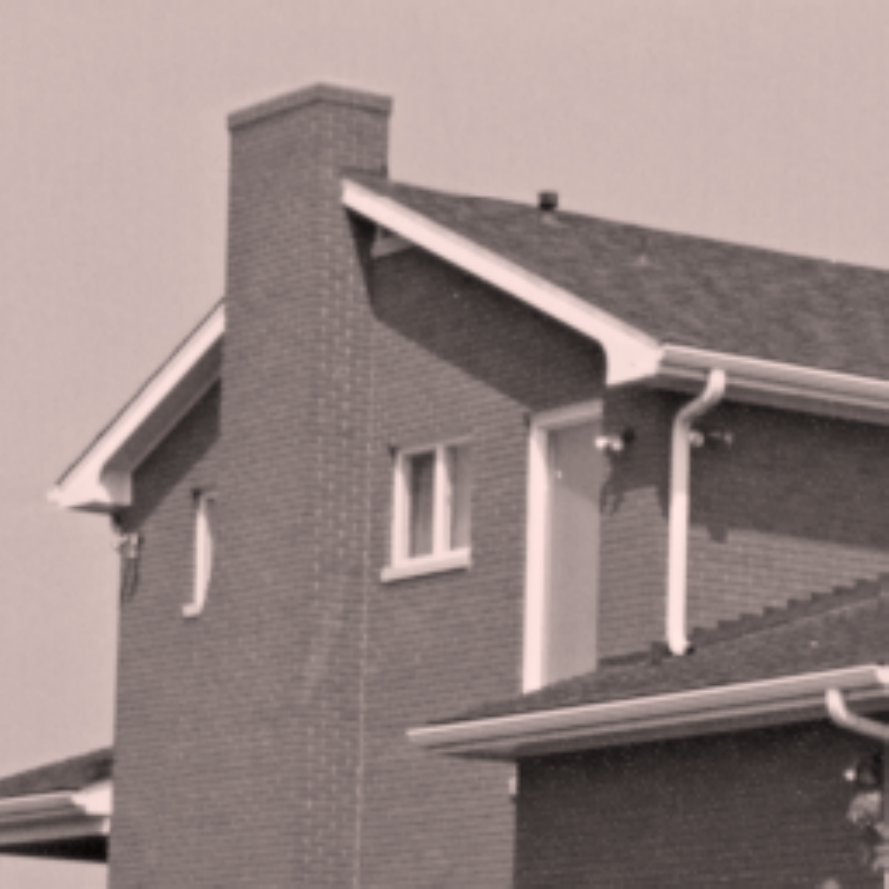}
\includegraphics[width=0.19\linewidth]{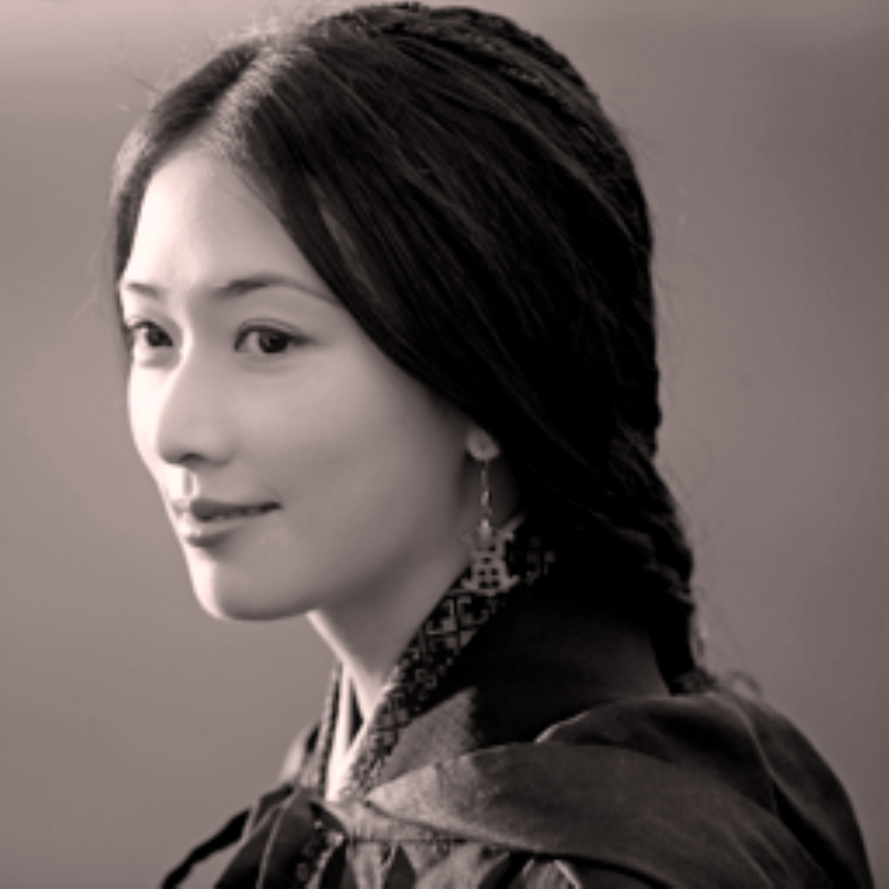}
\includegraphics[width=0.19\linewidth]{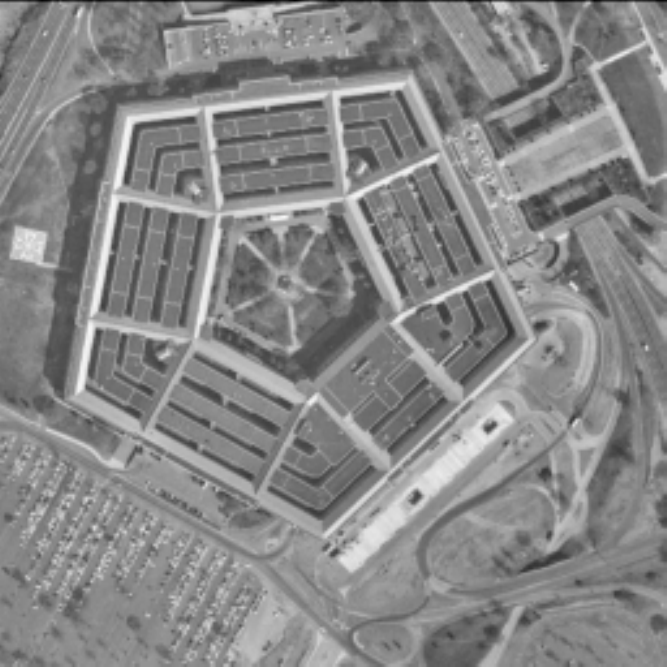}
\includegraphics[width=0.19\linewidth]{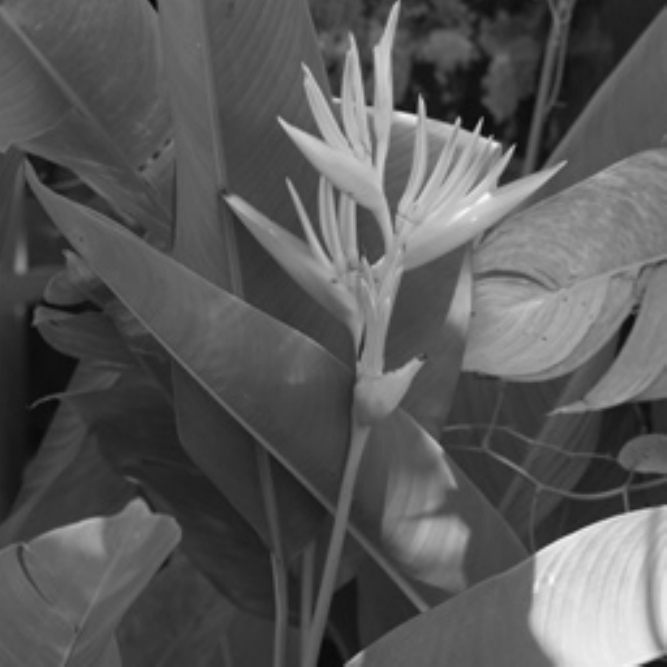}
\includegraphics[width=0.19\linewidth]{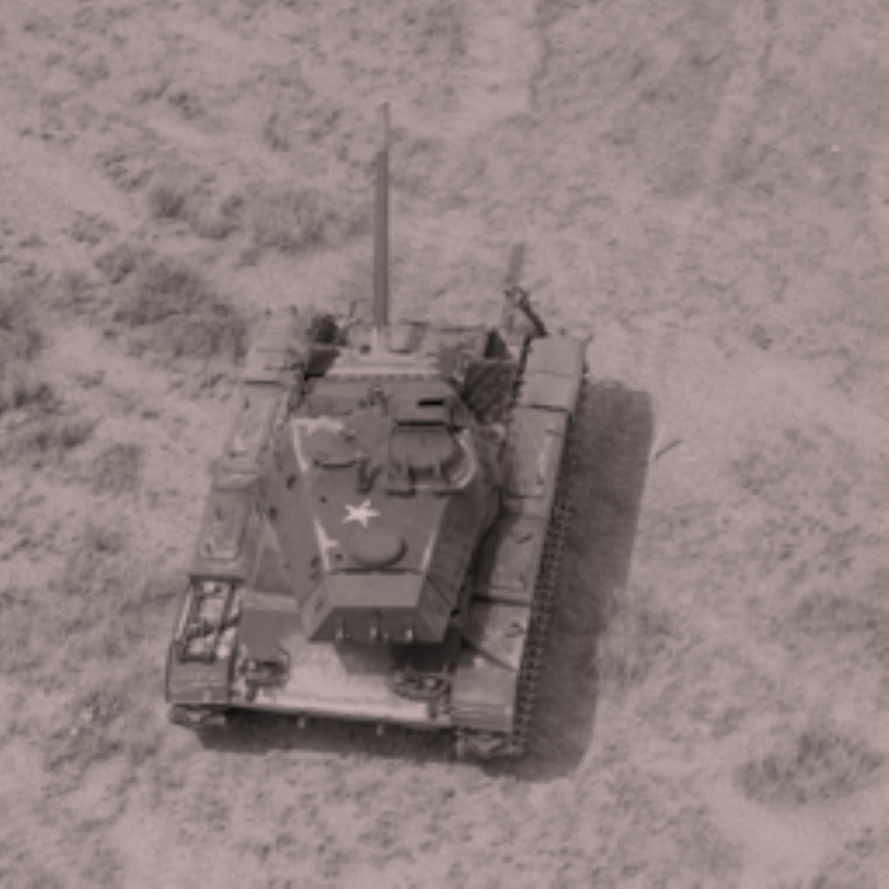}
\caption{Ten test images for performance evaluation of image denoising methods.}
\label{fig:4}
\end{figure}

We evaluate the competing methods on 10 widely used test images of size $256\times256$ shown in Fig.~\ref{fig:4}. To quantitatively evaluate the performance of the proposed method, the salt-and-pepper noise with various probability densities is added to those test images as the noisy observations. Typically, the results are shown on three noise levels, ranging from low noise level $P$= 10\%, to medium noise level $P$=30\% and to high noise level $P$= 50\%. Tables~\ref{tab:1} and \ref{tab:2} show these peak signal to noise ratio (PSNR) and structural similarity (SSIM) results of our denoising method compared with the competing denoising methods.

\begin{multicols}{2}
\begin{table*}[!htbp]
\renewcommand\arraystretch{1.33}
\caption{ Quantitative comparison of denoised  results with different  methods\protect\\  in terms of PSNR.}
%\centering  % ±í¾ÓÖÐ
%\tiny
%\vskip2mm
%\renewcommand{\baselinestretch}{0.9}
\setlength{\tabcolsep}{3.4pt}%Áпí
\begin{tabular}{cccccccc}
%\toprule
\hline
\multirow{2}{*}{\textbf{{}}}&\multirow{2}{*}{\textbf{{$P$}}}&\multirow{2}{*}{~\textbf{{PCP}}~}&\multirow{2}{*}{\textbf{{WNNM-}}}&\multirow{2}{*}{\textbf{{WSNM-}}}&\multirow{2}{*}{\textbf{{WSNM-}}}&\multirow{2}{*}{\textbf{{DWLP}}}&\multirow{2}{*}{\textbf{{DWLP}}}\\
& &{\textbf{{}}}& {\textbf{RPCA}} & {\textbf{RPCA}} &$\ell_1$ & \footnotesize{($p$=$q$=$1$)}&\\
\hline
{\textit{{}}}	  &$10\%$	  &26.540	  &27.636	  &27.974	  &26.452	  &34.595	  &35.026\\
{\textit{{}}}	  &$20\%$	  &25.906	  &26.114	  &26.340	  &25.080	  &30.957	  &31.571\\
{\textit{{couple}}}	  &$30\%$	  &24.946	  &24.374	  &24.540	  &23.907	  &28.026	  &28.771\\
{\textit{{}}}	  &$40\%$	  &23.418	  &23.723	  &23.774	  &22.621	  &26.107	  &26.573\\
{\textit{{}}}	  &$50\%$	  &18.468	  &23.233	  &23.332	  &21.065	  &22.894	  &22.946\\
 \hline							
{\textit{{}}}	  &$10\%$	  &30.815	  &31.924	  &30.917	  &30.163	  &40.570	  &40.473\\
{\textit{{}}}	  &$20\%$	  &29.336	  &28.636	  &29.877	  &28.636	  &36.629	  &36.381\\
{\textit{{elaine}}}	  &$30\%$	  &28.156	  &28.133	  &28.265	  &27.145	  &33.405	  &33.455\\
{\textit{{}}}	  &$40\%$	  &24.604	  &26.914	  &27.024	  &25.175	  &30.126	  &30.710\\
{\textit{{}}}	  &$50\%$	  &19.150	  &25.361	  &26.011	  &22.908	  &26.986	  &26.973\\
 \hline							
{\textit{{}}}	  &$10\%$	  &28.073	  &28.806	  &29.465	  &27.954	  &35.874	  &36.127\\
{\textit{{}}}	  &$20\%$	  &27.487	  &27.564	  &27.670	  &26.592	  &31.770	  &31.776\\
{\textit{{flower}}}	  &$30\%$	  &26.327	  &26.007	  &26.169	  &25.384	  &28.993	  &29.291\\
{\textit{{}}}	  &$40\%$	  &25.170	  &25.629	  &25.649	  &24.230	  &27.526	  &28.027\\
{\textit{{}}}	  &$50\%$	  &20.308	  &25.121	  &25.121	  &22.512	  &25.639	  &25.626\\
 \hline							
{\textit{{}}}	  &$10\%$	  &28.856	  &29.250	  &29.677	  &27.753	  &37.408	  &37.022\\
{\textit{{}}}	  &$20\%$	  &28.033	  &27.161	  &27.873	  &26.792	  &33.629	  &32.891\\
{\textit{{goldhill}}}	  &$30\%$	  &26.756	  &26.508	  &26.656	  &25.632	  &30.525	  &30.302\\
{\textit{{}}}	  &$40\%$	  &24.881	  &25.925	  &25.988	  &24.117	  &28.067	  &28.684\\
{\textit{{}}}	  &$50\%$	  &19.876	  &24.769	  &25.132	  &22.299	  &24.939	  &24.899\\
 \hline							
{\textit{{}}}	  &$10\%$	  &30.975	  &33.032	  &33.134	  &30.881	  &40.794	  &41.242\\
{\textit{{}}}	  &$20\%$	  &29.415	  &29.334	  &30.854	  &29.838	  &36.789	  &37.188\\
{\textit{{house}}}	  &$30\%$	  &28.342	  &29.274	  &29.426	  &28.150	  &33.284	  &33.893\\
{\textit{{}}}	  &$40\%$	  &24.713	  &27.685	  &27.791	  &25.971	  &30.109	  &31.038\\
{\textit{{}}}	  &$50\%$	  &18.212	  &25.074	  &26.071	  &23.184	  &25.566	  &25.623\\
 \hline							
{\textit{{}}}	  &$10\%$	  &28.727	  &30.587	  &30.785	  &28.339	  &35.554	  &35.815\\
{\textit{{}}}	  &$20\%$	  &28.018	  &28.968	  &29.382	  &26.722	  &33.034	  &33.044\\
{\textit{{lin}}}	  &$30\%$	  &27.352	  &27.703	  &27.847	  &26.184	  &30.836	  &30.583\\
{\textit{{}}}	  &$40\%$	  &25.487	  &26.145	  &26.296	  &23.714	  &27.990	  &27.226\\
{\textit{{}}}	  &$50\%$	  &20.125	  &24.858	  &25.070	  &21.248	  &25.486	  &25.510\\
 \hline							
{\textit{{}}}	  &$10\%$	  &24.590	  &27.894	  &28.322	  &26.726	  &33.349	  &33.366\\
{\textit{{}}}	  &$20\%$	  &23.697	  &26.008	  &26.401	  &25.160	  &29.945	  &29.547\\
{\textit{{monarch}}}	  &$30\%$	  &23.072	  &24.130	  &24.316	  &23.611	  &26.688	  &26.389\\
{\textit{{}}}	  &$40\%$	  &22.148	  &23.253	  &23.295	  &22.223	  &24.733	  &24.176\\
{\textit{{}}}	  &$50\%$	  &18.253	  &21.715	  &22.435	  &20.509	  &22.369	  &22.108\\
 \hline							
{\textit{{}}}	  &$10\%$	  &26.289	  &27.231	  &27.406	  &26.021	  &34.232	  &34.757\\
{\textit{{}}}	  &$20\%$	  &25.130	  &24.902	  &25.574	  &24.292	  &30.437	  &31.201\\
{\textit{{pentagon}}}	  &$30\%$	  &23.949	  &23.862	  &24.044	  &23.164	  &27.990	  &28.926\\
{\textit{{}}}	  &$40\%$	  &22.139	  &23.064	  &23.200	  &21.947	  &25.327	  &26.132\\
{\textit{{}}}	  &$50\%$	  &18.045	  &22.530	  &22.498	  &20.535	  &22.753	  &22.768\\
 \hline							
{\textit{{}}}	  &$10\%$	  &31.150	  &31.783	  &32.326	  &30.466	  &39.231	  &39.344\\
{\textit{{}}}	  &$20\%$	  &30.351	  &30.681	  &30.653	  &29.055	  &35.373	  &36.092\\
{\textit{{plants}}}	  &$30\%$	  &29.039	  &28.722	  &28.861	  &27.733	  &32.862	  &33.347\\
{\textit{{}}}	  &$40\%$	  &27.235	  &27.983	  &28.008	  &25.915	  &30.415	  &30.904\\
{\textit{{}}}	  &$50\%$	  &22.810	  &27.808	  &27.695	  &23.609	  &28.374	  &28.358\\
 \hline							
{\textit{{}}}	  &$10\%$	  &31.720	  &30.721	  &31.082	  &30.047	  &39.777	  &39.799\\
{\textit{{}}}	  &$20\%$	  &31.132	  &30.191	  &30.209	  &29.490	  &35.896	  &35.920\\
{\textit{{tank}}}	  &$30\%$	  &29.528	  &29.335	  &29.509	  &28.567	  &32.682	  &33.212\\
{\textit{{}}}	  &$40\%$	  &25.958	  &29.032	  &29.024	  &27.289	  &30.555	  &31.937\\
{\textit{{}}}	  &$50\%$	  &19.813	  &28.906	  &28.813	  &25.865	  &28.911	  &28.949\\
 \hline		
%\bottomrule
\end{tabular}
\label{tab:1}
\end{table*}
\end{multicols}

\begin{multicols}{2}
\begin{table*}[!htbp]
\renewcommand\arraystretch{1.33}
\caption{ Quantitative comparison of denoised results with different methods\protect \\ in terms of SSIM.}
%\centering  % ±í¾ÓÖÐ
%\tiny
%\vskip2mm
%\renewcommand{\baselinestretch}{1.9}
\setlength{\tabcolsep}{2.75pt}
\begin{tabular}{cccccccc}
%\toprule
\hline
\multirow{2}{*}{\textbf{{}}}&\multirow{2}{*}{\textbf{{$P$}}}&\multirow{2}{*}{~\textbf{{PCP}}~}&\multirow{2}{*}{\textbf{{WNNM-}}}&\multirow{2}{*}{\textbf{{WSNM-}}}&\multirow{2}{*}{\textbf{{WSNM-}}}&\multirow{2}{*}{\textbf{{DWLP}}}&\multirow{2}{*}{\textbf{{DWLP}}}\\
& &{\textbf{{}}}& {\textbf{RPCA}} & {\textbf{RPCA}} &$\ell_1$ & \footnotesize{($p$=$q$=$1$)}&\\
\hline
{\textit{{}}}	  &$10\%$	  &0.804	  &0.801	  &0.820	  &0.735	  &0.978	  &0.980\\
{\textit{{}}}	  &$20\%$	  &0.759	  &0.742	  &0.752	  &0.673	  &0.943	  &0.957\\
{\textit{{couple}}}	  &$30\%$	  &0.635	  &0.648	  &0.650	  &0.612	  &0.878	  &0.914\\
{\textit{{}}}	  &$40\%$	  &0.612	  &0.612	  &0.614	  &0.514	  &0.792	  &0.843\\
{\textit{{}}}	  &$50\%$	  &0.326	  &0.589	  &0.598	  &0.416	  &0.625	  &0.624\\
 \hline							
{\textit{{}}}	  &$10\%$	  &0.899	  &0.888	  &0.881	  &0.859	  &0.990	  &0.989\\
{\textit{{}}}	  &$20\%$	  &0.859	  &0.830	  &0.853	  &0.808	  &0.975	  &0.970\\
{\textit{{elaine}}}	  &$30\%$	  &0.804	  &0.820	  &0.823	  &0.755	  &0.951	  &0.948\\
{\textit{{}}}	  &$40\%$	  &0.623	  &0.795	  &0.797	  &0.689	  &0.897	  &0.920\\
{\textit{{}}}	  &$50\%$	  &0.321	  &0.745	  &0.766	  &0.565	  &0.809	  &0.808\\
 \hline							
{\textit{{}}}	  &$10\%$	  &0.821	  &0.819	  &0.839	  &0.776	  &0.976	  &0.977\\
{\textit{{}}}	  &$20\%$	  &0.786	  &0.782	  &0.785	  &0.743	  &0.939	  &0.945\\
{\textit{{flower}}}	  &$30\%$	  &0.707	  &0.721	  &0.724	  &0.666	  &0.876	  &0.905\\
{\textit{{}}}	  &$40\%$	  &0.661	  &0.705	  &0.705	  &0.617	  &0.810	  &0.854\\
{\textit{{}}}	  &$50\%$	  &0.362	  &0.684	  &0.690	  &0.500	  &0.729	  &0.727\\
 \hline							
{\textit{{}}}	  &$10\%$	  &0.824	  &0.800	  &0.812	  &0.744	  &0.980	  &0.978\\
{\textit{{}}}	  &$20\%$	  &0.781	  &0.739	  &0.754	  &0.699	  &0.949	  &0.941\\
{\textit{{goldhill}}}	  &$30\%$	  &0.683	  &0.697	  &0.698	  &0.631	  &0.896	  &0.895\\
{\textit{{}}}	  &$40\%$	  &0.646	  &0.678	  &0.676	  &0.558	  &0.807	  &0.846\\
{\textit{{}}}	  &$50\%$	  &0.393	  &0.631	  &0.641	  &0.464	  &0.673	  &0.669\\
 \hline							
{\textit{{}}}	  &$10\%$	  &0.896	  &0.889	  &0.887	  &0.851	  &0.989	  &0.990\\
{\textit{{}}}	  &$20\%$	  &0.861	  &0.833	  &0.865	  &0.825	  &0.973	  &0.978\\
{\textit{{house}}}	  &$30\%$	  &0.834	  &0.851	  &0.854	  &0.789	  &0.944	  &0.955\\
{\textit{{}}}	  &$40\%$	  &0.591	  &0.815	  &0.814	  &0.712	  &0.884	  &0.920\\
{\textit{{}}}	  &$50\%$	  &0.266	  &0.655	  &0.721	  &0.530	  &0.702	  &0.707\\
 \hline							
{\textit{{}}}	  &$10\%$	  &0.880	  &0.887	  &0.894	  &0.817	  &0.980	  &0.975\\
{\textit{{}}}	  &$20\%$	  &0.851	  &0.864	  &0.869	  &0.745	  &0.961	  &0.948\\
{\textit{{lin}}}	  &$30\%$	  &0.816	  &0.833	  &0.836	  &0.709	  &0.927	  &0.910\\
{\textit{{}}}	  &$40\%$	  &0.682	  &0.803	  &0.805	  &0.612	  &0.878	  &0.865\\
{\textit{{}}}	  &$50\%$	  &0.369	  &0.766	  &0.776	  &0.457	  &0.752	  &0.755\\
 \hline							
{\textit{{}}}	  &$10\%$	  &0.857	  &0.896	  &0.904	  &0.822	  &0.980	  &0.980\\
{\textit{{}}}	  &$20\%$	  &0.816	  &0.846	  &0.864	  &0.811	  &0.958	  &0.953\\
{\textit{{monarch}}}	  &$30\%$	  &0.787	  &0.803	  &0.806	  &0.732	  &0.914	  &0.917\\
{\textit{{}}}	  &$40\%$	  &0.684	  &0.776	  &0.777	  &0.664	  &0.867	  &0.879\\
{\textit{{}}}	  &$50\%$	  &0.422	  &0.687	  &0.733	  &0.561	  &0.754	  &0.776\\
 \hline							
{\textit{{}}}	  &$10\%$	  &0.770	  &0.771	  &0.798	  &0.728	  &0.969	  &0.975\\
{\textit{{}}}	  &$20\%$	  &0.716	  &0.691	  &0.707	  &0.627	  &0.923	  &0.943\\
{\textit{{pentagon}}}	  &$30\%$	  &0.589	  &0.601	  &0.604	  &0.560	  &0.859	  &0.903\\
{\textit{{}}}	  &$40\%$	  &0.590	  &0.570	  &0.573	  &0.488	  &0.738	  &0.810\\
{\textit{{}}}	  &$50\%$	  &0.358	  &0.554	  &0.557	  &0.401	  &0.590	  &0.586\\
 \hline							
{\textit{{}}}	  &$10\%$	  &0.900	  &0.866	  &0.876	  &0.836	  &0.987	  &0.989\\
{\textit{{}}}	  &$20\%$	  &0.848	  &0.843	  &0.845	  &0.782	  &0.968	  &0.975\\
{\textit{{plants}}}	  &$30\%$	  &0.779	  &0.795	  &0.798	  &0.719	  &0.935	  &0.943\\
{\textit{{}}}	  &$40\%$	  &0.728	  &0.779	  &0.779	  &0.655	  &0.882	  &0.904\\
{\textit{{}}}	  &$50\%$	  &0.440	  &0.775	  &0.775	  &0.507	  &0.809	  &0.808\\
 \hline							
{\textit{{}}}	  &$10\%$	  &0.825	  &0.764	  &0.773	  &0.722	  &0.978	  &0.980\\
{\textit{{}}}	  &$20\%$	  &0.788	  &0.748	  &0.749	  &0.699	  &0.943	  &0.953\\
{\textit{{tank}}}	  &$30\%$	  &0.711	  &0.725	  &0.730	  &0.659	  &0.883	  &0.912\\
{\textit{{}}}	  &$40\%$	  &0.558	  &0.716	  &0.716	  &0.603	  &0.791	  &0.863\\
{\textit{{}}}	  &$50\%$	  &0.257	  &0.711	  &0.711	  &0.538	  &0.679	  &0.689\\
\hline
%\bottomrule
\end{tabular}
\label{tab:2}
\end{table*}
\end{multicols}
Here, we have the following observations. First, the origional PCP method has a poor performance when the image is heavily corrupted by salt-and-pepper noise, since the dense noise destroys sparse and low-rank priors; Second, the proposed DWLP method achieves the highest PSNR and SSIM consistently on all the five noise levels. It achieves 7.03dB-9.26dB, 5.10dB-8.74dB, 4.75dB-9.40dB, 6.64dB-10.36dB and 0.81dB-1.60dB improvements over PCP, WNNM-RPCA, WSNM-RPCA, WSNM-$\ell_1$ and DWLP($p$=$q$=$1$) under 10\% salt-and-pepper noise level, and 1.57dB-4.45dB, 1.757dB-4.929dB, 1.634dB-4.72dB, 2.78dB-6.31dB and 0.1dB-1.34dB under 30\% salt-and-pepper noise level, respectively; Third, as the noise level increases, some methods such as DWLP($p$=$q$=$1$) may get a higher PSNR than DWLP, but our DWLP method achieves a higher SSIM, indicating a more satisfactory visual performance.

\begin{figure*}[!htpb]
\centering
\subfigure[ ]{\label{fig:5 subfig:a}
\includegraphics[width=0.23\linewidth]{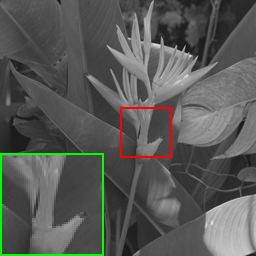}}
\hspace{0.001pt}
\subfigure[ ]{\label{fig:5 subfig:b}
\includegraphics[width=0.23\linewidth]{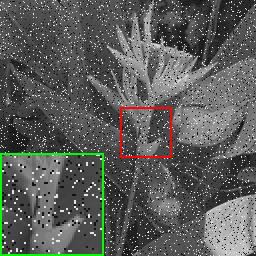}}
\hspace{0.001pt}
\subfigure[ ]{\label{fig:5 subfig:c}
\includegraphics[width=0.23\linewidth]{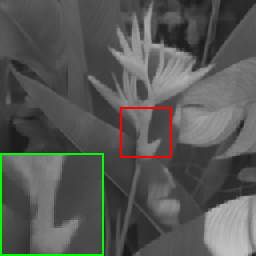}}
\hspace{0.001pt}
\subfigure[ ]{\label{fig:5 subfig:d}
\includegraphics[width=0.23\linewidth]{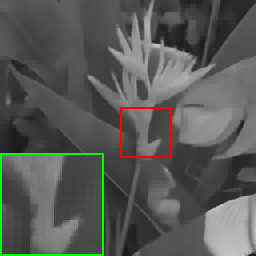}}
\vfill
\hspace{0.001pt}
\subfigure[ ]{\label{fig:5 subfig:e}
\includegraphics[width=0.23\linewidth]{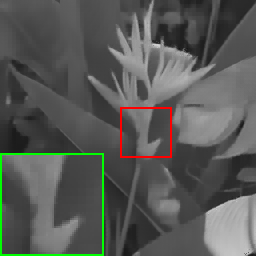}}
\hspace{0.001pt}
\subfigure[ ]{\label{fig:5 subfig:f}
\includegraphics[width=0.23\linewidth]{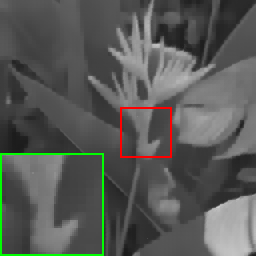}}
\hspace{0.001pt}
\subfigure[ ]{\label{fig:5 subfig:g}
\includegraphics[width=0.23\linewidth]{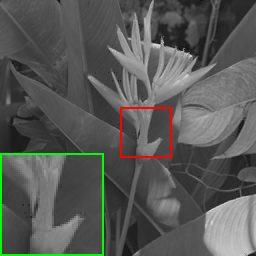}}
\hspace{0.001pt}
\subfigure[ ]{\label{fig:5 subfig:h}
\includegraphics[width=0.23\linewidth]{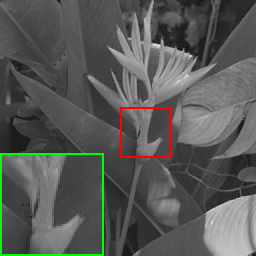}}
\caption{Denoised results on image \textit{plants} by different methods (noise level 10\%). (a) Ground Truth; (b) Noisy image (16.791dB/0.208); (c)  PCP \cite{pcp} (31.150dB/0.900); (d) WNNM-RPCA \cite{wnnmGu,wsnmrpcaXie} (31.783dB/0.866); (e) WSNM-RPCA \cite{wsnmrpcaXie} (32.326dB/0.876); (f) WSNM-$\ell_p$ \cite{Chen} (30.466dB/0.836); (g) DWLP ($p$=$q$=$1$) \cite{ddwPeng} (39.231dB/0.987); (h) DWLP (39.344dB/0.989).}
\label{fig:5}
\end{figure*}

\begin{figure*}[!htpb]
\centering
\subfigure[ ]{\label{fig:6 subfig:a}
\includegraphics[width=0.23\linewidth]{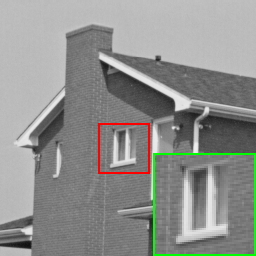}}
\hspace{0.001pt}
\subfigure[ ]{\label{fig:6 subfig:b}
\includegraphics[width=0.23\linewidth]{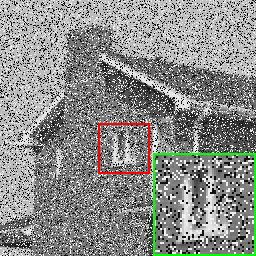}}
\hspace{0.001pt}
\subfigure[ ]{\label{fig:6 subfig:c}
\includegraphics[width=0.23\linewidth]{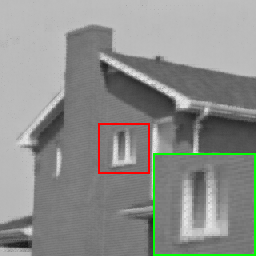}}
\hspace{0.001pt}
\subfigure[ ]{\label{fig:6 subfig:d}
\includegraphics[width=0.23\linewidth]{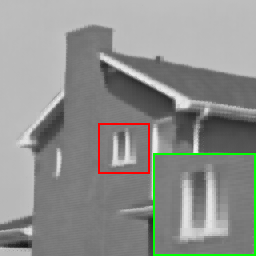}}
\vfill
\hspace{0.001pt}
\subfigure[ ]{\label{fig:6 subfig:e}
\includegraphics[width=0.23\linewidth]{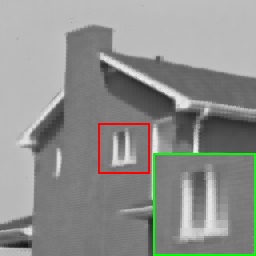}}
\hspace{0.001pt}
\subfigure[ ]{\label{fig:6 subfig:f}
\includegraphics[width=0.23\linewidth]{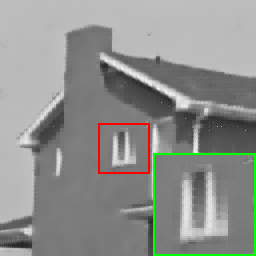}}
\hspace{0.001pt}
\subfigure[ ]{\label{fig:6 subfig:g}
\includegraphics[width=0.23\linewidth]{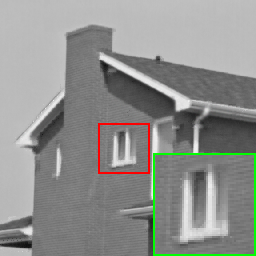}}
\hspace{0.001pt}
\subfigure[ ]{\label{fig:6 subfig:h}
\includegraphics[width=0.23\linewidth]{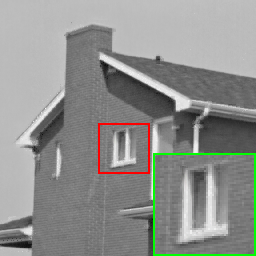}}
\caption{Denoised results on image \textit{house} by different methods (noise level 10\%). (a) Ground Truth; (b) Noisy image (11.725dB/0.740); (c)  PCP \cite{pcp} (28.342dB/0.834); (d) WNNM-RPCA \cite{wnnmGu,wsnmrpcaXie} (29.274dB/0.851); (e) WSNM-RPCA \cite{wsnmrpcaXie} (29.426dB/0.854); (f) WSNM-$\ell_p$ \cite{Chen} (28.150dB/0.789); (g) DWLP ($p$=$q$=$1$) \cite{ddwPeng} (33.284dB/0.944); (h) DWLP (33.893dB/0.955).}
\label{fig:6}
\end{figure*}

\begin{figure*}[!htpb]
\centering
\subfigure[ ]{\label{fig:7 subfig:a}
\includegraphics[width=0.23\linewidth]{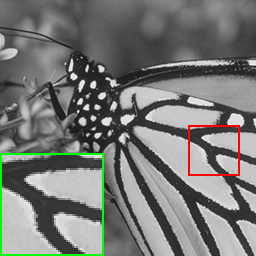}}
\hspace{0.001pt}
\subfigure[ ]{\label{fig:7 subfig:b}
\includegraphics[width=0.23\linewidth]{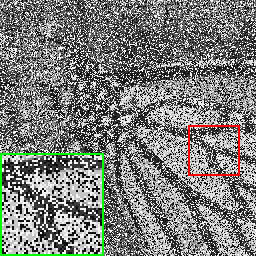}}
\hspace{0.001pt}
\subfigure[ ]{\label{fig:7 subfig:c}
\includegraphics[width=0.23\linewidth]{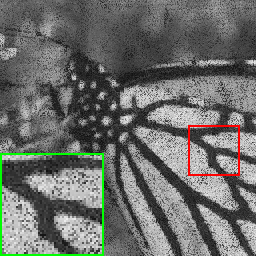}}
\hspace{0.001pt}
\subfigure[ ]{\label{fig:7 subfig:d}
\includegraphics[width=0.23\linewidth]{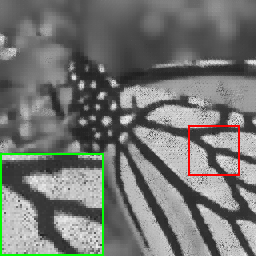}}
\vfill
\hspace{0.001pt}
\subfigure[ ]{\label{fig:7 subfig:e}
\includegraphics[width=0.23\linewidth]{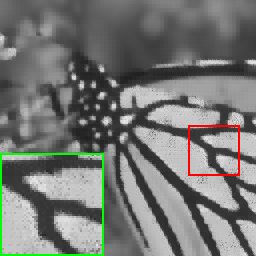}}
\hspace{0.001pt}
\subfigure[ ]{\label{fig:7 subfig:f}
\includegraphics[width=0.23\linewidth]{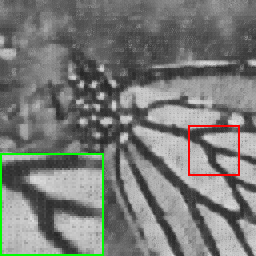}}
\hspace{0.001pt}
\subfigure[ ]{\label{fig:7 subfig:g}
\includegraphics[width=0.23\linewidth]{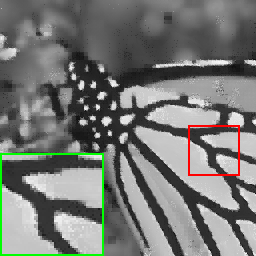}}
\hspace{0.001pt}
\subfigure[ ]{\label{fig:7 subfig:h}
\includegraphics[width=0.23\linewidth]{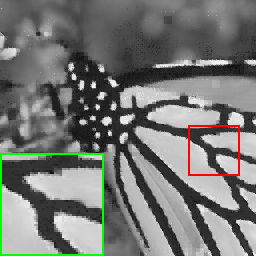}}
\caption{Denoised results on image \textit{monarch} by different methods (noise level 10\%). (a) Ground Truth; (b) Noisy image (9.077dB/0.080); (c)  PCP \cite{pcp} (18.253dB/0.422); (d) WNNM-RPCA \cite{wnnmGu,wsnmrpcaXie} (21.715dB/0.687); (e) WSNM-RPCA \cite{wsnmrpcaXie} (22.435dB/0.733); (f) WSNM-$\ell_p$ \cite{Chen} (20.509dB/0.561); (g) DWLP ($p$=$q$=$1$) \cite{ddwPeng} (22.369dB/0.754); (h) DWLP (22.108dB/0.776).}
\label{fig:7}
\end{figure*}

Then, we compare the visual quality of the denoised images by the competing methods under different noise levels. The ground truth images of size 256$\times$256 pixels are shown in Figs.~\ref{fig:5 subfig:a}, \ref{fig:6 subfig:a} and \ref{fig:7 subfig:a}. The noisy images corrupted by 10\% salt-and-pepper noise, 30\% salt-and-pepper noise and 50\% salt-and-pepper noise are shown in Fig.~\ref{fig:5 subfig:b}, Fig.~\ref{fig:6 subfig:b} and Fig.~\ref{fig:7 subfig:b}. The recovered images obtained by PCP \cite{pcp}, WNNM-RPCA \cite{wnnmGu,wsnmrpcaXie}, WSNM-RPCA \cite{wsnmrpcaXie}, WSNM-$\ell_1$ \cite{Chen}, DWLP($p$=$q$=$1$) \cite{ddwPeng} and DWLP are shown in Figs.~\ref{fig:5},~\ref{fig:6} and~\ref{fig:7}(c)-(h) respectively. The regions of size 50$\times$50 marked with red boxes are shown in zoom-in windows to have a close-up observation data.

According to the recovery experiments of the image \textit{plants} with 10\% salt-and-pepper noise in Fig.~\ref{fig:5}, we observed that much more appealing results are obtained by our DWLP. One can clearly discern the plant branches recovered by DWLP and DWLP($p$=$q$=$1$) as shown in the zoom-in windows of Figs.~\ref{fig:5 subfig:g} and \ref{fig:5 subfig:h}, which are hard to be distinguished in the other methods. Since the DWLP and DWLP($p$=$q$=$1$) are able to enhance low-rank and sparsity simultaneously for matrix recovery that helps to capture data structure. Compared with DWLP($p$=$q$=$1$), our method not only obtains a higher PSNR and SSIM, but also preserves more details, because of the weaker power and the balanced manner. When the noise level is increased to 30\%, it can be seen in the zoom-in window of Fig.~\ref{fig:6} that our method can better preserve the brick structure of the wall, while the other methods tend to over-smooth the textures or have some noise points remained. Although the PSNR of DWLP is lower than that of DWLP($p$=$q$=$1$) with 50\% salt-and-pepper noise, our DWLP obtains a higher SSIM and better preserve wing veins of the butterfly, as shown in Figs.~\ref{fig:7 subfig:g} and \ref{fig:7 subfig:h}.

\subsection{Discussions}
\label{5.2}
\begin{figure}[!htpb]
\centering
\includegraphics[scale=0.55]{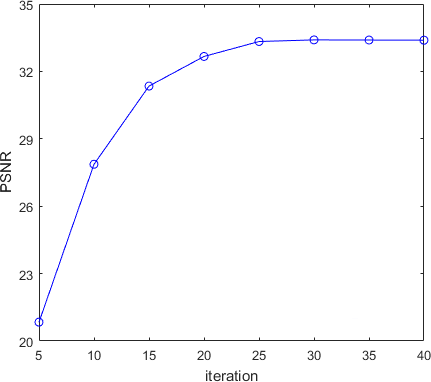}
\caption{The PSNR increment during the iteration.}
\label{fig:8}
\end{figure}

First of all, we discuss the convergence of the proposed DWLP method. Since the DWLP model is nonconvex, it is difficult to provide its theoretical proof of global convergence. Hereby, we verify the convergence of DWLP through an experiment in terms of the PSNR increments as the evaluation criterion. We present the denoising performance gains from each iteration based on the 10 test images added with $30\%$ salt-and-pepper noise. In Fig.~\ref{fig:8}, the average PSNR increments of the 10 test images tend to be unchangeable after a small number of iterations, accounting for the convergence of the proposed DWLP method.

Next, we discuss the setting of the parameters in experiments. Some parameters can be readily fixed by experiences: for the $256\times256 $ test images, the step size is set to 4 (denotes the neighborhood image patches are extracted in every 4 pixels), and the patch size $h$ is 8. According to the convergence analysis in Fig.~\ref{fig:8}, the number of iterations is set to 30.

\begin{figure}[!htpb]
\centering
\subfigure[ ]{\label{fig:8 subfig:a}
\includegraphics[width=0.7\linewidth]{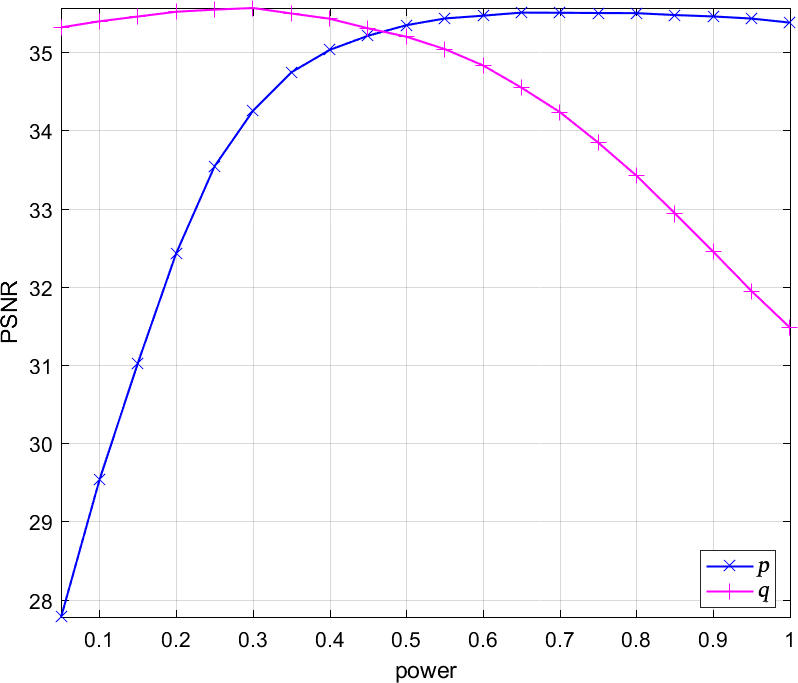}}
\hspace{0.001\linewidth}
\subfigure[ ]{\label{fig:8 subfig:b}
\includegraphics[width=0.7\linewidth]{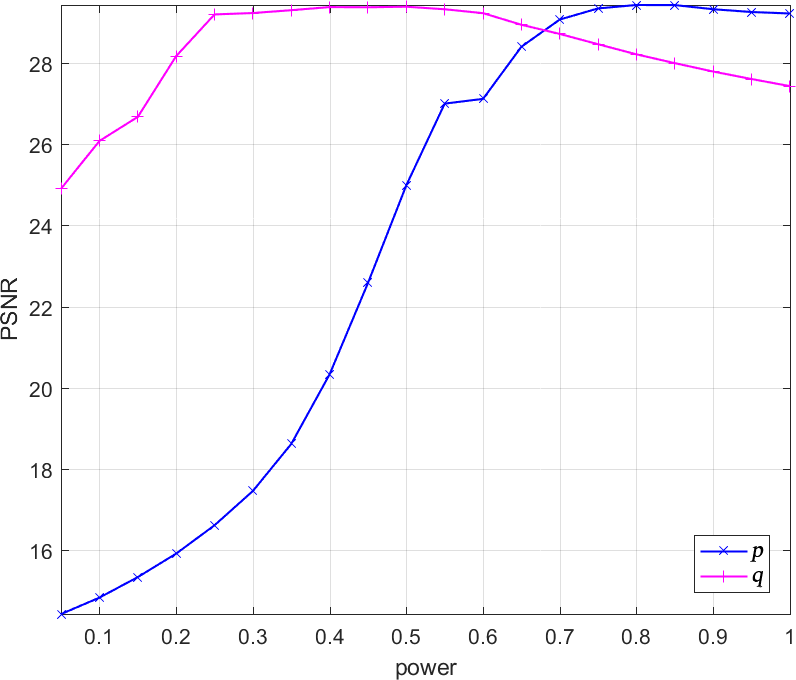}}
\caption{PSNR values with different $p$ and $q$ on both low and medium noise levels. (a) low noise level ($10\%$); (b) medium noise level ($30\%$).}
\label{fig:9}
\end{figure}

\begin{figure}[!htpb]
\centering
\subfigure[ ]{\label{fig:9 subfig:a}
\includegraphics[width=0.7\linewidth]{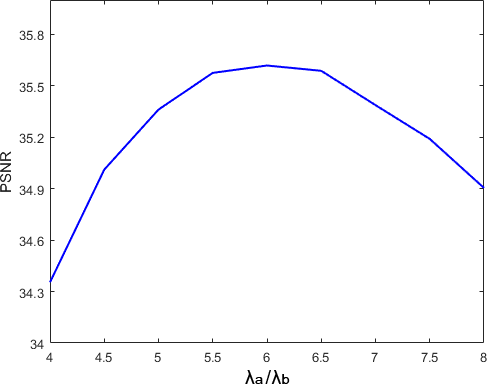}}
\hspace{0.001\linewidth}
\subfigure[ ]{\label{fig:9 subfig:b}
\includegraphics[width=0.7\linewidth]{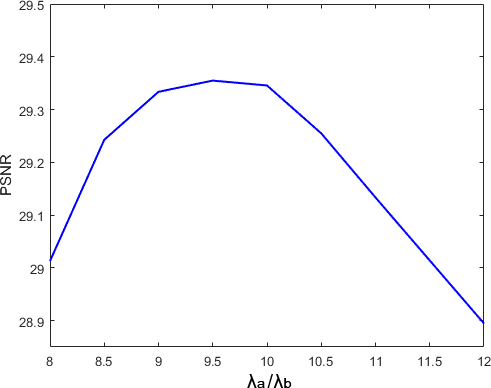}}
\caption{PSNR values with different $\lambda_a /{\lambda_e} $ on both low and medium noise levels. (a) low noise level ($10\%$); (b) medium noise level ($30\%$).}
\label{fig:10}
\end{figure}

Then, we analyze the setting of the other parameters ($K$, $\lambda_a $, $\lambda_e $ and powers $p $, $q $) through several experiments. Different from the empirical value based on the observation that more than 80\% of image patches recur 9 times or more in the original image scale \cite{nss1}, the number of similar patches $K$ is suggested to be $h^2 $ in our method. The parameter experiments are based on the test images, and the average PSNR and SSIM results are listed in Table~\ref{tab:3}. Since it can be observed from Table~\ref{tab:3} that the increments of PSNR and SSIM become quite small when more than 64 similar patches are used, we set $K$ as 64 to avoid unnecessary computation. The setting $K=h^2=64$ helps enhance the low-rank constraint as a square matrix. As for the norm powers $p$ and $q$, we determine their optimal values by investigating the influence of their difference value upon the quality of restored results. In this test, the powers $p$ and $q$ are uniformly sampled range from 0.1 to 1 with an interval 0.1, and Fig.~\ref{fig:9}~~shows the PSNR of denoised results with different powers under both low and medium noise levels. Moreover, we find that DWLP is less sensitive to either $\lambda_a $ or $\lambda_e $ than their ratio $\lambda_a /{\lambda_e} $, for the optimal values of $\lambda_a /{\lambda_e}$ can guarantee to achieve a perfect balance between enforcing low-rank and separating sparse noise. Fixing other optimization parameters, we analyze the PSNR versus different values of the trade-off parameter ratio $\lambda_a /{\lambda_e} $ in Fig.~\ref{fig:10}, and note that the ratio is increasing with noise density. The DWLP achieves the highest PSNR at $\lambda_a /{\lambda_e} $= 6.358 under the low noise level, and $\lambda_a /{\lambda_e} $= 10 under the medium noise level, Table~\ref{tab:4} summarizes all the suggested values of $p$, $q$ and $\lambda_a /{\lambda_e}$ under different noise level.

\begin{multicols}{2}
\begin{table*}[!htbp]
\renewcommand\arraystretch{1.8}
\caption{PSNR and SSIM of the DWLP with different values of $K$ on medium \protect \\ noise level}
%\scriptsize
%\centering  % ±í¾ÓÖÐ
\setlength{\tabcolsep}{4.4pt}
\begin{tabular}{ccccccccc}
 \hline
 \multirow{1}{*}{$K$}        & 8 & 22 & 36 & 50 & 64 & 78 & 92\\
 \hline
 \multirow{1}{*}{PSNR}       & 6.372 & 11.524 & 21.993 & 28.722 & 29.372 & 29.605 & 29.490\\
 \multirow{1}{*}{SSIM}       & 0.112 & 0.506 & 0.803 & 0.891 & 0.904 & 0.906 & 0.900\\
 \hline
\end{tabular}
\label{tab:3}
\end{table*}
\end{multicols}

\begin{multicols}{2}
\begin{table*}[!htbp]
\renewcommand\arraystretch{1.6}
\caption{The optimal parameter values of the DWLP on different noise levels}
%\scriptsize
%\centering  % ±í¾ÓÖÐ
\setlength{\tabcolsep}{9pt}
\begin{tabular}{ccccccccc}
 \hline
 \multirow{1}{*}{}                          & $10\%$ & $20\%$ & $30\%$ & $40\%$ & $50\%$\\
 \hline
 \multirow{1}{*}{$p$}                      & 0.650 & 0.765 & 0.800 & 0.905 & 0.916\\
 \multirow{1}{*}{$q$}                      & 0.340 & 0.393 & 0.419 & 0.570 & 0.595 \\
 \multirow{1}{*}{$\lambda_a /{\lambda_e} $} & 6.358 & 7.738 & 10.003 & 10.792 & 13.866 \\
 \hline
\end{tabular}
\label{tab:4}
\end{table*}
\end{multicols}

\section{Conclusion}
\label{6}
In this paper, we propose the DWLP model to improve the performance of the RPCA method, which integrates both the weighted method and $\ell_p$ quasi-norm into the low-rank regularization term and the sparsity regularization term respectively. The $p$-shrinkage is introduced to solve the $\ell_p$-norm minimization, which has more stability in the nonconvex optimization and provides a more accurate estimation for the low-rank and sparse matrix recovery problem. The DWLP makes impressively quantitative improvements not only on the low-rank approximation of NSS matrices to deal with the overshrink problem, but also on the image denoising experiments. It preserves much better image structures and details with less visual artifacts on visual quality, outperforming the original PCP optimization, the WNNM-RPCA, the WSNM-RPCA and the DWLP($p$=$q$=$1$) greatly. In light of promise of DWLP, deeper investigations of our method remain: combining with other restoration theory to the developed method, such as dictionary representation, a more effective method is considered as one of the future work.

\protect \quad\\

\protect \quad\\

\protect \quad\\

\protect \quad\\

\protect \quad\\

\protect \quad\\

\protect \quad\\

\protect \quad\\

\protect \quad\\

\protect \quad\\

\begin{IEEEbiography}[{\includegraphics[width=1in,height=1.25in,clip,keepaspectratio]{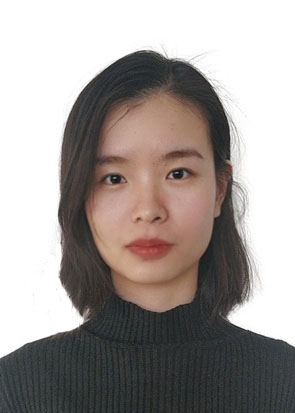}}]{Dong Huiwen}  Master student at the Faculty of Information Technology, Beijing University of Technology. Her research interest is image processing.
\end{IEEEbiography}
\begin{IEEEbiography}[{\includegraphics[width=1in,height=1.25in,clip,keepaspectratio]{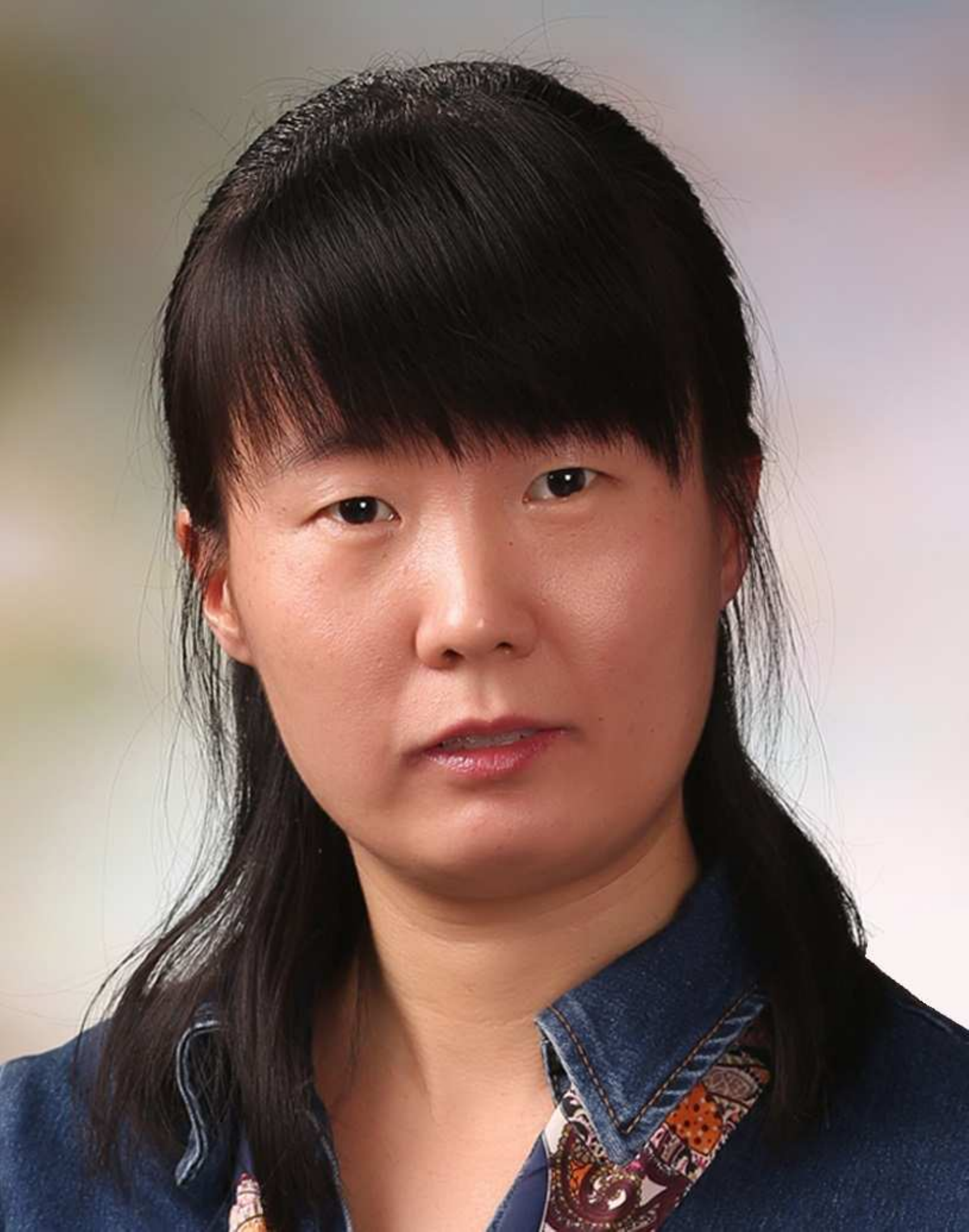}}]{YU Jing} Lecturer and MA student supervisor at the Faculty of Information Technology, Beijing University of Technology. She received her Ph.D. degree from Tsinghua University in 2011. Her research interest covers image processing and pattern recognition.
\end{IEEEbiography}
\begin{IEEEbiography}[{\includegraphics[width=1in,height=1.25in,clip,keepaspectratio]{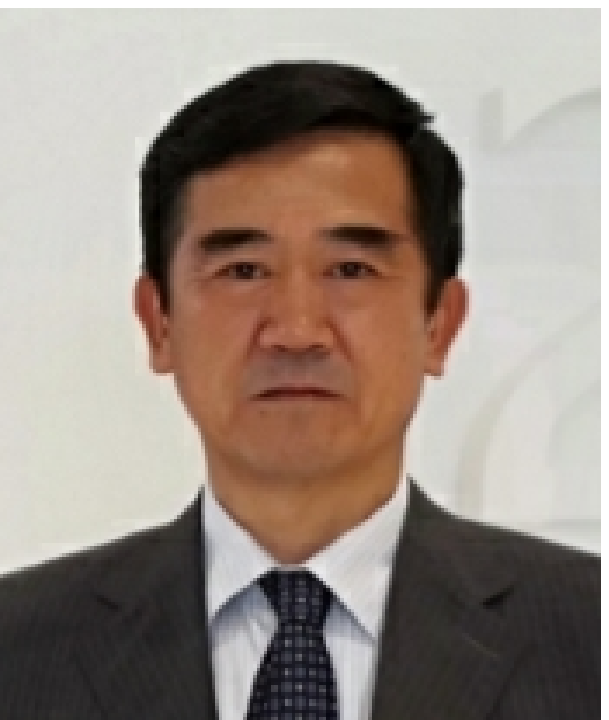}}]{XIAO Chuangbai}Professor and Ph. D. supervisor at the Faculty of Information Technology, Beijing University of Technology. His research interest covers digital signal processing, audio and video signal processing, and network communication. Corresponding author of this paper.
\end{IEEEbiography}
\EOD

\protect \quad\\

\protect \quad\\

\protect \quad\\

\protect \quad\\

\protect \quad\\

\protect \quad\\

\protect \quad\\

\protect \quad\\

\protect \quad\\


\subsection{References}
\begin{thebibliography}{00}

\bibitem{rpcaJohn}%C
J.~Wright, A.~Ganesh, S.~Rao, Y.~Peng, Y.~Ma,
Robust principal component analysis: Exact recovery of corrupted low-rank matrices via convex optimization,
 presented at NIPS. Neural Information Processing Systems, 2009, pp. 2080-2088,
[Online] Available: http://papers.nips.cc.

\bibitem{rpcaCandes}
E.~J. Cand\'{e}s, X.~Li, Y.~Ma, J.~Wright,
"Robust principal component analysis?"
 Journal of the ACM. J. ACM., vol. 58, no. 3, pp. 11:1-11:37, May. 2011. %DOI: 10.1145/1970392.1970395.

\bibitem{rpcaJi}
H.~Ji, S.~Huang, Z.~Shen, Y.~Xu,
"Robust video restoration by joint sparse and low rank matrix approximation,"
 SIAM Journal on Imaging Sciences, vol. 4, no. 4, pp. 1122-1142, Nov. 2011. %DOI: 10.1137/100817206.

\bibitem{video1}
S.~E. Ebadi, E.~Izquierdo, S.~E. Ebadi, E.~Izquierdo,
"Foreground segmentation with tree-structured sparse rpca,"
 IEEE Transactions on Pattern Analysis and Machine Intelligence, vol. 40, no. 9, pp. 2273-2280, Sept. 2018. %DOI: 10.1109/TPAMI.2017.2745573.

\bibitem{ddwPeng}
Y.~Peng, J.~Suo, Q.~Dai, W.~Xu,
"Reweighted low-rank matrix recovery and its application in image restoration,"
 IEEE Transactions on Cybernetics, vol. 44, no. 12, pp. 2418-2430, Mar. 2014. %DOI: 10.1109/TCYB.2014.2307854.

\bibitem{face1}
B.~Jiang, K.~Jia,
"Robust facial expression recognition algorithm based on local metric learning,"
 Journal of Electronic Imaging, vol. 25, no. 1, pp. 013022, Feb. 2016.

\bibitem{recommend1}
T.~Dai, T.~Gao, Z.~Li, X.~Cai, S.~Pan,
"Low-rank and sparse matrix factorization for scientific paper recommendation in heterogeneous network,"
 IEEE Access, vol. 6, pp. 59015-59030, Aug. 2018. %DOI: 10.1109/ACCESS.2018.2865115.

\bibitem{recommend2}%C
Z.~L. Zhao, L.~Huang, C.~D. Wang, J.~H. Lai, P.~S. Yu,
Low-rank and sparse matrix completion for recommendation,
 presented at ICONIP. International Conference on Neural Information Processing, 2017, pp. 3-13,
[Online] Available: https://doi.org/10.1007/978-3-319-70139-4-1.

\bibitem{recommend3}%C
J.~Feng, H.~Xu, S.~Yan,
Online Robust PCA via Stochastic Optimization,
 presented at NIPS. Neural Information Processing Systems, 2013, pp. 404-412,
[Online] Available: http://papers.nips.cc.

\bibitem{sto}
D.~Donoho,
"De-noising by soft threshold,"
 IEEE Transactions on Information Theory, vol. 41, no. 3, pp. 613-627, May. 1995.

\bibitem{wl1mCandes}
E.~J. Cand{\`e}s, M.~B. Wakin, S.~P. Boyd,
"Enhancing sparsity by reweighted $\ell_1$ minimization,"
 Journal of Fourier Analysis and Applications, vol. 14, no. 5, pp. 877-905, Dec. 2008.

\bibitem{wnnmGu}%C
S.~Gu, L.~Zhang, W.~Zuo, X.~Feng,
Weighted nuclear norm minimization with application to image denoising,
 presented at CVPR. The IEEE conference on computer vision and pattern recognition, 2014, pp. 2862-2869.

\bibitem{weightsXu}%C
W.~Xu, M.~A. Khajehnejad, S.~Avestimehr, B.~Hassibi,
Breaking through the thresholds: an analysis for iterative reweighted $\ell_1$ minimization via the grassmann angle framework,
 presented at the 2010 IEEE International Conference on Acoustics, Speech and Signal Processing, 2010, pp. 5498-5501.

\bibitem{lp1}
H.~Liu, R.~Xiong, D.~Liu, S.~Ma, F.~Wu, W.~Gao,
"Image denoising via low rank regularization exploiting intra and inter patch correlation,"
 IEEE Transactions on Circuits and Systems for Video Technology, vol. 28, no. 12, pp. 3321-3332, Dec. 2018.

\bibitem{lp2}
M.~Wang, J.~Yu, J.-H. Xue, W.~Sun,
"Denoising of hyperspectral images using group low-rank representation,"
 IEEE Journal of Selected Topics in Applied Earth Observations and Remote Sensing, vol. 9, no. 9, pp. 4420-4427, Sept. 2016.

\bibitem{lp3}
L.~Liu, W.~Huang, D.-R. Chen,
"Exact minimum rank approximation via schatten $p$-norm minimization,"
 Journal of Computational and Applied Mathematics, vol. 267, pp. 218-227, Sept. 2014.

\bibitem{lp4}%C
F.~Nie, H.~Huang, C.~Ding,
Low-rank matrix recovery via efficient schatten $p$-norm minimization,
 presented at Twenty-Sixth AAAI Conference on Artificial Intelligence, 2012.

\bibitem{wsnmXie}
Y.~Xie, S.~Gu, Y.~Liu, W.~Zuo, W.~Zhang, L.~Zhang,
"Weighted schatten $p$-norm minimization for image denoising and background subtraction,"
 IEEE Transactions on Image Processing, vol. 25, no. 10, pp. 4842-4857, Aug. 2016.

\bibitem{wsnmrpcaXie}
Y.~Xie, Y.~Qu, D.~Tao, W.~Wu, Q.~Yuan, W.~Zhang,
"Hyperspectral image restoration via iteratively regularized weighted schatten $p$-norm minimization,"
 IEEE Transactions on Geoscience and Remote Sensing, vol. 54, no. 8, pp. 4642-4659, Aug. 2016.

\bibitem{Chen}
Chen,~G., Wang,~J., Zhang,~F. and Wang,~W.,
"Image denoising in impulsive noise via weighted Schatten p-norm regularization."
 Journal of Electronic Imaging, vol. 28, no. 1, Jan. 2019.

\bibitem{reviewer3_2}%C
C. Zhao, J. Zhang, S. Ma and W. Gao,
Nonconvex $\ell_p$ Nuclear Norm based ADMM Framework for Compressed Sensing,
 present at 2016 Data Compression Conference (DCC), 2016, pp. 161-170.

 \bibitem{reviewer3_1}
Z.~Zhang, M.~Zhao, F.~Li, L.~Zhang, S.~Yan,
"Robust Alternating Low-Rank Representation by joint $\ell_p$- and $\ell_{2,p}$-norm minimization,"
 Neural Networks, vol. 96, pp. 55-70, Dec. 2017.

\bibitem{Chartrand2}
J.~Woodworth and R.~Chartrand,
"Compressed sensing recovery via nonconvex shrinkage penalties,"
 Inverse Problems, vol. 32, pp. 75004¨C75028, May. 2016.

\bibitem{gst}%C
W.~Zuo, D.~Meng, L.~Zhang, X.~Feng and D.~Zhang, %Sydney, NSW
A Generalized Iterated Shrinkage Algorithm for Non-convex Sparse Coding,
 present at 2013 IEEE International Conference on Computer Vision, 2013, pp. 217-224.

\bibitem{Chartrand3}%C
R. Chartrand,
Fast algorithms for nonconvex compressive sensing: MRI reconstruction from very few data,
 present at 2009 IEEE International Symposium on Biomedical Imaging: From Nano to Macro, 2009, pp. 262-265.

\bibitem{Chartrand1}
R.~Chartrand,
"Nonconvex Splitting for Regularized Low-Rank + Sparse Decomposition,"
 IEEE Transaction on Signal Proessing, vol. 60, no. 11, pp. 218-227, Nov. 2012.

\bibitem{peppersalt1}%C
G.~Liu, P.~Li,
Recovery of coherent data via low-rank dictionary pursuit,
 presented at NIPS. Neural Information Processing Systems, 2014, pp. 1206-1214,
[Online] Available: http://papers.nips.cc.

\bibitem{med}
C.~Khare, K.~K. Nagwanshi,
"Image restoration technique with non linear filter,"
 International Journal of Advanced Science and Technology, vol. 39, pp. 67-74, Jan. 2012.

\bibitem{wmf}
H.~C. Bandala~Hern\'{a}ndez, J.~Rocha-P\'{e}rez, A.~Diaz-sanchez, J.~Lemus-L\'{o}pez, H.~V\'{a}zquez-Leal, A.~Diaz-Armendariz, J.~Ramirez-Angulo,
"Weighted median filters: An analog implementation,"
 Integration, the VLSI Journal, vol. 55, pp. 227-231, Sep. 2016.

\bibitem{cwmf1}
S.~Yazdi, F.~Homayouni,
"Modified adaptive center weighted median filter for supperssing impulse noise in image,"
 Int J Res Rev Appl Sci, vol. 1, no. 3, pp. 37-45, Aug. 2009.

\bibitem{cwmf2}
T.~Chen, H.~R. Wu,
"Adaptive impulse detection using center-weighted median filters,"
 IEEE Signal Processing Letters, vol. 8, no. 1, pp. 1-3, Jan. 2001.

\bibitem{amf1}
T.~Loupas, W.~N. McDicken, P.~L. Allan,
"An adaptive weighted median filter for speckle suppression in medical ultrasonic images,"
 IEEE Transactions on Circuits and Systems, vol. 36, no. 1, pp. 129-135, Jan. 1989.

\bibitem{amf2}%arXiv
S.~Shrestha,
"Image denoising using new adaptive based median filter,"
 arXiv preprint arXiv: 1410.2175, 2014.

\bibitem{dba}
K.~S. Srinivasan, D.~Ebenezer,
"A new fast and efficient decision-based algorithm for removal of high-density impulse noises,"
 IEEE Signal Processing Letters, vol. 14, no. 3, pp. 189-192, Feb. 2007.

\bibitem{idba}
R.~N. KULKARNI, P.~C. Prof.~BHASKAR,
"Implementation of decision based algorithm for median filter to extract impulse noise,"
 Int. J. Adv. Res. Electr. Electron. Instrum. Eng., vol. 2, no. 6, pp. 2507-2512, Aug. 2013.

\bibitem{pdba}
C.~A. V.~S. Balasubramanian, G., \textit{et al}.,
"Probabilistic decision based filter to remove impulse noise using patch else trimmed median,"
 AEU-International Journal of Electronics and Communications, vol. 70, no. 4, pp. 471-481, Apr. 2016.

\bibitem{variational}%book
M~. Bertalm¨ªo,
"Denoising of Photographic Images and Video Fundamentals, Open Challenges and New Trends,"
 present at Computer Vision and Pattern Recognition. USA, 2018.

\bibitem{rof}
L.I.~Rudin, S.~Osher, E.~Fatemi,
"Nonlinear total variation based noise removal algorithms,"
 Phys. D Nonlinear Phenom, vol. 60, pp. 259-268, Nov. 1992.

\bibitem{4_TV}%C
M.~Sakurai, S.~Kiriyama, T.~Goto, S.~Hirano,
Fast algorithm for total variation minimization,
 present at 2011 18th IEEE International Conference on Image Processing (ICIP), 2011, pp. 1461¨C1464.

\bibitem{ogstv1}
I. W. Selesnick, Farshchian, "Sparse Signal Approximation via Nonseparable Regularization,"
 IEEE Transactions on Signal Processing, vol. 65, no. 10, pp. 2561-2575, May. 2017.

\bibitem{ogstv2}%C
I. W. Selesnick, P. Chen,
Total variation denoising with overlapping group sparsity,
 present at 2013 IEEE International Conference on Acoustics, 2013, pp. 5696-5700.

\bibitem{tvsolution}
R. Chartrand, B. Wohlberg,
Total-variation regularization with bound constraints,
 present at 2010 IEEE International Conference on Acoustics, 2010, pp. 766-769.

\bibitem{bm3d1}
I.~Djurovi{\'{c}},
"Bm3d filter in salt-and-pepper noise removal,"
 EURASIP Journal on Image and Video Processing, vol. 2016, no. 1, pp. 13, Mar. 2016.

\bibitem{bm3d2}
K.~Dabov, A.~Foi, V.~Katkovnik, K.~Egiazarian,
"Image denoising by sparse 3-d transform-domain collaborative filtering,"
 IEEE Transactions on Image Processing, vol. 16, no. 8, pp. 2080-2095, July. 2007. %DOI: 10.1109/TIP.2007.901238

\bibitem{lr1}
F.~Lam, D.~Liu, Z.~Song, N.~Schuff, Z.~P. Liang,
"A fast algorithm for denoising magnitude diffusion-weighted images with rank and edge constraints,"
 Magnetic Resonance in Medicine, vol. 75, no. 1, pp. 433-440, Mar. 2015.

\bibitem{lr2}%C
Z.~Zha, X.~Yuan, B.~Wen, J.~Zhou, C.~Zhu,
Joint patch-group based sparse representation for image inpainting,
 present at Asian Conference on Machine Learning, 2018, pp. 145-160.

\bibitem{lr3}
Q.~Guo, S.~Gao, X.~Zhang, Y.~Yin, C.~Zhang,
"Patch-based image inpainting via two-stage low rank approximation,"
 IEEE transactions on visualization and computer graphics, vol. 24, no. 6, pp. 2023-2036, May. 2018.

\bibitem{sparseCoding}%C
Z.~Gu, M.~Shao, L.~Li, Y.~Fu,
Discriminative metric: Schatten norm vs. vector norm,
 present at the 21st International Conference on Pattern Recognition (ICPR2012), 2012, pp. 1213-1216.

\bibitem{Zha}%C
Z.~Zha, X.~Liu, X.~Huang, H.~Shi, Y.~Xu, Q.~Wang, L.~Tang, X.~Zhang,
Analyzing the group sparsity based on the rank minimization methods,
 present at 2017 IEEE International Conference on Multimedia and Expo (ICME), 2017, pp. 883-888.

\bibitem{Candes}
E.~J. Cand\'{e}s, B.~Recht,
"Exact matrix completion via convex optimization,"
 Foundations of Computational Mathematics, vol. 9, no. 6, pp. 717-772, Apr. 2009.

\bibitem{pcp}%C
Z.~Zhou, X.~Li, J.~Wright, E.~Candes, L.~Yu,
Stable principal component pursuit,
 present at 2010 IEEE International Symposium on Information Theory, 2010, pp. 1518-1522.

\bibitem{aboutnnm}%C
P.~L. Guangcan~Liu,
Recovery of coherent data via low-rank dictionary pursuit,
 present at Advances in Nueral Information Processing Systems (NIPS), 2014, pp. 1206-1214.

\bibitem{svt}%C
A.~Beck, M.~Teboulle,
A fast iterative shrinkage-thresholding algorithm with application to wavelet-based image deblurring,
 present at 2009 IEEE International Conference on Acoustics, Speech and Signal Processing, 2009, pp. 693-696.

\bibitem{tnnr}%C
D.~Zhang, Y.~Hu, J.~Ye, X.~Li, X.~He,
Matrix completion by truncated nuclear norm regularization,
 present at 2012 IEEE Conference on Computer Vision and Pattern Recognition, 2012, pp. 2192-2199.

\bibitem{psm}%C
T.~Oh, H.~Kim, Y.~Tai, J.~Bazin, I.~S. Kweon,
Partial sum minimization of singular values in rpca for low-level vision,
 present at 2013 IEEE International Conference on Computer Vision, 2013, pp. 145-152.

\bibitem{imgres}
L.~Liu, W.~Huang, D.-R. Chen,
"Exact minimum rank approximation via schatten $p$-norm minimization,"
Journal of Computational and Applied Mathematics, vol. 267, pp. 218-227, Sept. 2014.

\bibitem{gsvt}%C
C.~Lu, C.~Zhu, C.~Xu, S.~Yan, Z.~Lin,%(2015, February).
Generalized singular value thresholding,
 present at Twenty-Ninth AAAI Conference on Artificial Intelligence, 2015, pp. 1805-1811.

\bibitem{nss2}%C
Z.~Zha, X.~Zhang, Q.~Wang, Y.~Bai, L.~Tang,
Image denoising using group sparsity residual and external nonlocal self-similarity prior
 present at 2017 IEEE International Conference on Image Processing (ICIP), 2017, pp. 2956-2960.

\bibitem{alm}%arXiv
Z.~Lin, M.~Chen, Y.~Ma,
"The augmented lagrange multiplier method for exact recovery of corrupted low-rank matrices,"
 arXiv preprint arXiv:1009.5055, 2010.

\bibitem{Zhang}
X.~{Zhang}, J.~{Zheng}, Y.~{Yan}, L.~{Zhao}, R.~{Jiang},
"Joint Weighted Tensor Schatten $p$-Norm and Tensor $\ell_p $-Norm Minimization for Image Denoising,"
 Journal of IEEE Access, vol. 7, pp. 20273-20280, Feb. 2019.

\bibitem{Zha2}
Z.~{Zha}, X.~{Yuan},
"Bridge the Gap Between Group Sparse Coding and Rank Minimization via Adaptive Dictionary Learning"
 arXiv preprint arXiv:1709.03979, 2017.

\bibitem{nss1}%C
D.~Glasner, S.~Bagon, M.~Irani,
Super-resolution from a single image,
 present at 2009 IEEE 12th International Conference on Computer Vision, 2009, pp. 349-356.

\bibitem{prefilterMatching}%C
H.~{Ji}, C.~{Liu}, Z.~{Shen}, Y.~{Xu},
Robust video denoising using low rank matrix completion,
 present at 2010 IEEE Computer Society Conference on Computer Vision and Pattern Recognition, 2010, pp. 1791-1798.


%\bibitem{Chartrand4}%C
%R.~Chartrand,
%Shrinkage mappings and their induced penalty functions,
%present at 2014 IEEE International Conference on Acoustics, Speech and Signal Processing (ICASSP), 2014, pp. 1026-1029.

%\bibitem{fr_TV}
%Z.~Ren, C.~He, Q.~Zhang,
%"Nonlinear total variation based noise removal algorithms,"
%Signal Processing, vol. 93, pp. 2408-2421, Sept. 2013.

%\bibitem{tvsolution2}
%MLA Salvadeo, Denis H. P. , N. D. A. Mascarenhas, A. L. M. Levada, "Nonlocal Markovian models for image denoising." Journal of Electronic Imaging, vol. 25, no. 1, pp. 013003, Jan. 2017.
\end{thebibliography}
\end{document}